\documentclass{article}
\usepackage{authblk}
\usepackage{placeins}

\usepackage{caption}
\usepackage{array} 

\usepackage{graphicx} 
\usepackage{booktabs} 

\usepackage{float} 

\usepackage{multirow}

\usepackage{amsmath,amsthm,amsfonts,amssymb,amscd,pifont}
\usepackage{verbatim}
\usepackage{url}

\usepackage{makecell}

\usepackage{mathrsfs}
\usepackage[ansinew]{inputenc}
\usepackage{graphicx}
\usepackage{a4wide}
\usepackage{subfigure}
\usepackage{graphics}

\usepackage{tabularx}
\usepackage{booktabs}

\usepackage{graphicx}
\usepackage{hyperref}
\usepackage{xfrac}
\usepackage{array}
\usepackage[all]{xy}
\usepackage{epic}
\usepackage{latexsym}
\usepackage{enumerate}
\usepackage{appendix}

\newtheorem*{Theorem*}{Theorem}

\newtheorem{maintheorem}{Theorem}

\newcommand{\cmt}{\begin{maintheorem}}
\newcommand{\fmt}{\end{maintheorem}}

\newtheorem{maincorollary}[maintheorem]{Corollary}

\newcommand{\cmc}{\begin{maincorollary}}
\newcommand{\fmc}{\end{maincorollary}}

\newtheorem{T}{Theorem}[section]
\newcommand{\ct}{\begin{T}}
\newcommand{\ft}{\end{T}}
\newtheorem{D}{Definition}[section]
\newcommand{\cd}{\begin{D}}
\newcommand{\fd}{\end{D}}

\newtheorem{Corollary}[T]{Corollary}
\newcommand{\cco}{\begin{Corollary}}
\newcommand{\fco}{\end{Corollary}}

\newtheorem{Proposition}[T]{Proposition}
\newcommand{\cpr}{\begin{Proposition}}
\newcommand{\fpr}{\end{Proposition}}

\newtheorem{Lemma}[T]{Lemma}
\newcommand{\cle}{\begin{Lemma}}
\newcommand{\fle}{\end{Lemma}}

\newtheorem{Sublemma}[T]{Sublemma}
\newcommand{\csle}{\begin{Sublemma}}
\newcommand{\fsle}{\end{Sublemma}}

\usepackage{graphicx}
\usepackage{xcolor}
\usepackage{mathtools}
\usepackage{mathrsfs}

\newcommand{\ra}{\rightarrow}

\newcommand{\ep}{\epsilon}

\newcommand{\nn}{\nonumber}
\newcommand{\be}{\begin{eqnarray}}
\newcommand{\ee}{\end{eqnarray}}

\newtheorem{thm}{Theorem}

\newcommand{\R}{\mathbb{R}}
\newcommand{\C}{\mathbb{C}}

\newcommand{\N}{\mathbb{N}}

\usepackage{tikz}
\usetikzlibrary{positioning, shapes.geometric}

\usepackage{pgfplots}
\usepackage{pgfplotstable}
\usepackage{tikz}
\usetikzlibrary{shapes.geometric, arrows.meta, positioning}

\usepackage{pgfplots}
\usepgfplotslibrary{polar}
\pgfplotsset{compat=1.18}

\tikzstyle{block} = [rectangle, draw, fill=blue!20, 
    text width=8em, text centered, rounded corners, minimum height=3em]
\tikzstyle{line} = [draw, ->, thick]  

\title{Cauchy activation function and XNet}

\author[1,2]{Xin Li\thanks{Email: \texttt{xinli2023@u.northwestern.edu}}}
\author[3,4]{Zhihong Xia\thanks{Corresponding author: \texttt{xia@math.northwestern.edu}}}
\author[3,5]{Hongkun Zhang\thanks{Email: \texttt{hzhang@umass.edu}}}

\affil[1]{Department of Computer Science, Northwestern University, Evanston, IL, USA}

\affil[2]{Mathematical Modelling and Data Analytics Center, Oxford
  Suzhou Centre for Advanced Research, Suzhou, China}

\affil[3]{School of Natural Science, Great Bay University, Guangdong, China}
\affil[4]{Department of Mathematics, Northwestern University, Evanston, IL, USA}

\affil[5]{Department of Mathematics and Statistics, University of Massachusetts Amherst, Amherst, MA, USA}

\date{\today}

\begin{document}

\maketitle

\begin{abstract}

  We have developed a novel activation function, named the {\em Cauchy
  Activation Function}. This function is derived from the {\em Cauchy
  Integral Theorem} in complex analysis and is specifically tailored
  for problems requiring high precision. This innovation has led to
  the creation of a new class of neural networks, which we call
  (Comple)XNet, or simply XNet.

  We will demonstrate that XNet is particularly effective for
  high-dimensional challenges such as image classification and solving
  Partial Differential Equations (PDEs). Our evaluations show that
  XNet significantly outperforms established benchmarks like MNIST and
  CIFAR-10 in computer vision, and offers substantial advantages over
  Physics-Informed Neural Networks (PINNs) in both low-dimensional and
  high-dimensional PDE scenarios.




\end{abstract}

\section{Introduction}

In today's scientific exploration, the rise of computational
technology has marked a significant turning point. Traditional methods
of theory and experimentation are now complemented by advanced
computational tools that tackle the complexity of real-world
systems. Machine learning, particularly deep neural networks, has led
to breakthroughs in fields like image processing and language
understanding \cite{Goodfellow2016, LeCun2015}, and its application to
scientific problems--such as predicting protein structures
\cite{Senior2020, Jumper2021} or forecasting weather
\cite{Reichstein2019}--demonstrates its potential to revolutionize our
approach.

One of the primary challenges in computational mathematics and
artificial intelligence (AI) lies in determining the most appropriate
function to accurately model a given dataset. In machine learning, the
objective is to leverage such functions for predictive
purposes. Traditional methods rely on predetermined classes of
functions, such as polynomials or Fourier series, which, though simple
and computationally manageable, may limit the flexibility and accuracy
of the fit. In contrast, modern deep learning neural networks
primarily employ locally linear functions with nonlinear activations.

While the trend in deep learning has been towards increasingly deeper architectures (from 8-layer AlexNet to 152/1001-layer ResNets), our work demonstrates an alternative approach. The Cauchy activation function's superior approximation capabilities allow us to achieve comparable or better performance with significantly simplified architectures. For example, in our MNIST experiments, we reduced three fully connected layers to a single layer while maintaining high accuracy. This suggests that the power of neural networks may not solely lie in depth, but also in the choice of activation functions that can capture complex patterns more efficiently.

This finding has important implications for both theoretical understanding and practical applications:

1. Computational Efficiency: Fewer layers mean reduced computational costs and memory requirements

2. Training Stability: Shallower networks are typically easier to train and less prone to vanishing gradient problems

3. Interpretability: Simpler architectures may be more interpretable than very deep networks

\subsection{Algorithm Development}

In our previous work \cite{ZLX24}, we introduced the initial concept
of extending real-valued functions into the complex domain, using the
Cauchy integral formula to device a machine learning algorithm, and
demonstrating its high accuracy and performance through examples in
time series. In this work, we introduce a more general method stemming
from the same mathematical principles. This approach is not confined
to addressing mathematical physics problems such as low-dimensional
PDEs; it also effectively tackles a broad spectrum of AI application
issues, including image processing. This paper primarily showcases
examples in image processing and both low-dimensional and
high-dimensional (100-dimensional) PDEs. We are actively continuing
our research to explore further applications, with very promising
preliminary results.

Recent advancements in deep learning for solving PDEs and CV
capabilities are well-documented, with significant contributions from
transformative architectures that integrate neural networks with PDEs
to enhance function approximation in high-dimensional spaces
\cite{Sirignano2018, Raissi2019, Jin2021, Transolver2024,
  PINNsFormer2023}. In CV, comprehensive surveys and innovative
methods have significantly advanced visual processing techniques
\cite{le2022deep, kaur2023comprehensive, wu2020visual,
  fu2024featup}. 

Also, we have extensively reviewed literature on integrating complex
domains into neural network architectures. Explorations of
complex-valued neural networks and further developments highlight the
potential of complex domain methodologies in modern neural network
frameworks \cite{Hirose2012, Lee2022, Barrachina2023, Begion2016,
  Begion2018, Yeats2021}.

Feedforward Neural Networks (FNNs), despite their capabilities, are
often limited by the granularity of approximation they can achieve due
to traditional activation functions such as ReLU or Sigmoid
\cite{Jarrett2009, Hinton2010, Rumelhart1986}. Recent works aim to
optimize network performance through the discovery of more effective
activation functions, leading to significant advancements in network
functionality and computational efficiency \cite{Boulle2020, Kunc2024,
  ModSwish2024, Dubey2022, Bingham2022, Ramachandran2017,
  Knezevic2023, Tripathi2021}. Adaptive activation functions have demonstrated significant improvements in convergence rates and accuracy, particularly in deep learning and physics-informed neural networks \cite{Jagtap2020a, Jagtap2020b, Jagtap2022}. A recent survey \cite{Jagtap2022Survey} highlights their critical role in enhancing neural network performance across various tasks.

We propose a novel method that does not directly incorporate complex
numbers. Drawing on insights from the complex Cauchy integral formula,
we utilize Cauchy kernels as basis functions and introduce a new
Cauchy activation function. This innovative approach significantly
enhances the network's precision and efficacy across various tasks,
particularly in high-dimensional spaces.

\begin{figure}[h!]
    \centering
    \begin{minipage}[b]{0.4\textwidth}
        \centering
        \includegraphics[width=\linewidth]{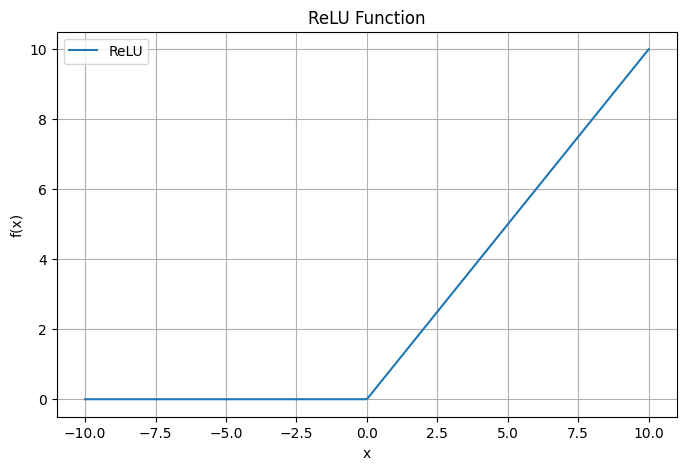}
        \caption{ReLU }
        \label{fig:reluscatter}
    \end{minipage}\hfill
    \begin{minipage}[b]{0.4\textwidth}
        \centering
        \includegraphics[width=\linewidth]{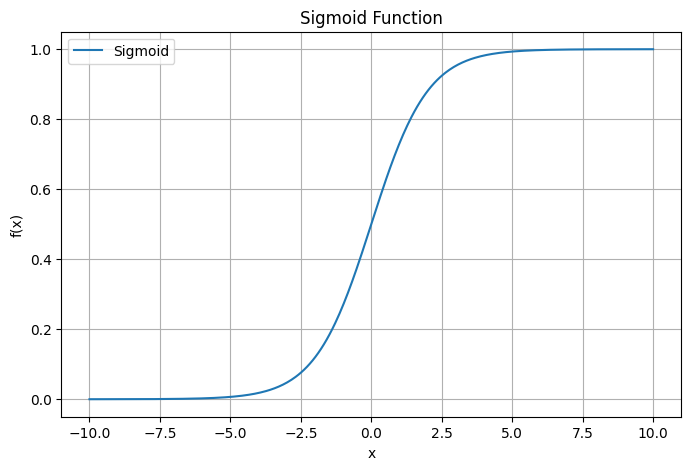}
        \caption{Sigmoid }
        \label{fig:sigmoidscatter}
    \end{minipage}
\end{figure}

The Cauchy activation function can be expressed as follows 
$$\phi_{\lambda_1, \lambda_2, d}(x) = \frac{\lambda_1 * x}{x^2+d^2}+  \frac{\lambda_2}{x^2+d^2},$$ where $\lambda_1, \lambda_2, d$ are trainable parameters.


Theoretically, the Cauchy activation function can approximate any
smooth function to its highest possible order. Moreover, from the
figure below, we observe that our activation function is localized and
decays at the both ends. This feature turned out to be very useful in
approximating local data. This capability to finely tune to specific
data segments sets it apart significantly from traditional activation
functions such as ReLU.

\begin{figure}[h!]
    \centering
    \begin{minipage}[b]{0.7\textwidth}
        \centering
        \includegraphics[width=\linewidth]{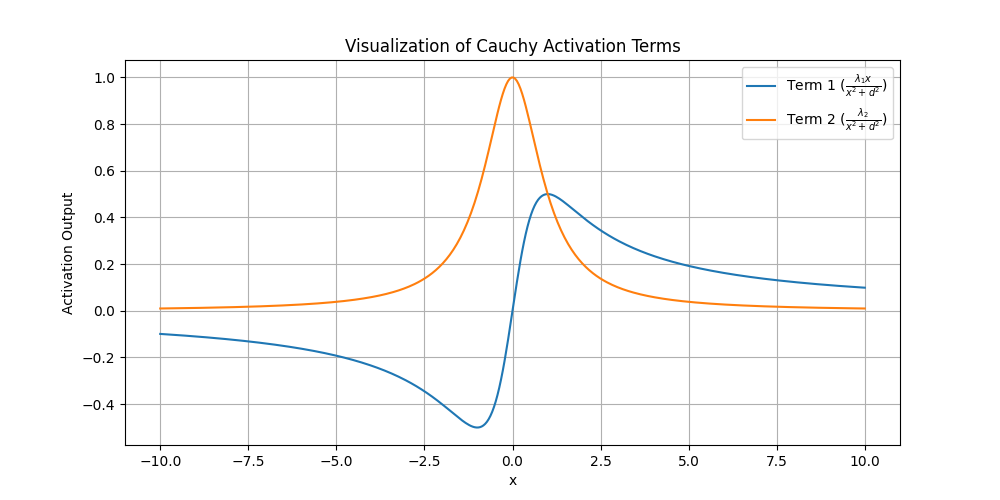}
        \caption{2 terms of Cauchy activation, with $\lambda_1=\lambda_2=d=1$}
        \label{fig:acticauchy}
    \end{minipage}\hfill
   \end{figure}

In Section \ref{sec:cauchy}, we will explain how the Cauchy activation
function is derived and why it is mathematically efficient.

\subsection{Enhanced Neural Network Efficiency with Cauchy Activation Function: High-Order Approximation and Beyond}

Recent advancements in machine learning have been pivotal in tackling complex scientific challenges, such as high-dimensional Partial Differential Equations (PDEs) and nonlinear systems \cite{Sirignano2018, Raissi2019, Transolver2024}. Traditional neural networks that employ activation functions like ReLU or Sigmoid often necessitate deep, multi-layered architectures to adequately approximate complex functions. This requirement typically results in increased computational complexity and extensive parameter demands.

In contrast, our innovative approach utilizes a \textit{single-layer network} equipped with the Cauchy activation function, derived directly from our Cauchy Approximation Theorem (Theorem \ref{thm:CAT}). This methodology introduces a fundamentally different and more efficient approximation mechanism:
\begin{equation}
    \left|f - \sum_{j=1}^m \frac{\lambda_j}{(\xi_1^j - z_1)\cdots(\xi_N^j - z_N)}\right|_{L^\infty(M)} \leq C m^{-k},
\end{equation}
where \( m \) represents the number of network parameters. This approximation exhibits an \( O(m^{-k}) \) convergence rate for any \( k > 0 \), a substantial improvement over traditional ReLU networks, which typically achieve \( O(m^{-r/d}) \) convergence for \( r \)-smooth functions. Here, d represents the input dimensionality, and r denotes the smoothness degree of the function, as detailed in \cite{yarotsky2017error, e2018exponential}.

Given a desired approximation error \( \epsilon > 0 \), the comparative network sizes required to achieve this error are significantly reduced:
\begin{equation}
    m_{\text{ReLU}} = O\left((1/\epsilon)^{d/r}\right) \quad \text{versus} \quad m_{\text{Cauchy}} = O\left((1/\epsilon)^{1/k}\right),
\end{equation}
where \( d \) and \( r \) denote the input dimensionality and smoothness of the function, respectively, highlighting the drastic reduction in complexity that the Cauchy activation function enables.

This theoretical advantage has been translated directly into empirical performance improvements. Detailed examples and results of these experiments are discussed in Section \ref{sec:application}. 

In our MNIST experiments, even a simple two-layer network [128, 64] equipped with the Cauchy activation function outperformed a similar ReLU-based network, achieving a validation accuracy of 96.71\% compared to 96.22\%. Remarkably, a single-layer network [100] equipped with the Cauchy activation function achieved a competitive 96.30\% accuracy while requiring significantly fewer epochs to converge compared to traditional activations.

The effectiveness of the Cauchy activation function extends across a range of tasks, from image classification, where it achieved 91.42\% accuracy on CIFAR-10, outperforming ReLU's 90.91\%, to high-dimensional PDE solving, where it reduced error from 0.0349 to 0.00354. 

\section{Approximation Theorems}
\label{sec:cauchy}

Consider, for example, a dataset comprising values of certain function
$g(x_i), i =1, \ldots, n$ corresponding to points $x_1, \ldots, x_n$
on the real line. These data may contain noises. We begin by assuming that the
target function for fitting is a real-analytic function $f$. Although
this assumption may appear stringent, it is important to note that
non-analytic functions, due to their unpredictable nature and form,
exhibit a weak dependence on specific data sets. Another perspective
is that our aim is to identify the most suitable analytic function
that best fits the provided dataset.

Real analytic functions can be extended to complex plane. The central
idea of our algorithm is to place observers in the complex
plane. Similar to activation for each node in artificial neural
network, a weight is computed and assigned to each observer. The value
of the predicted function $f$ at any point is then set to be certain
weighted average of all observers. Our core mathematical theory is the
Cauchy Approximation Theorem (Section \ref{sec:cauchy}, Theorem
\ref{thm:CAT}), derived in the next section. The Cauchy Approximation
Theorem guarantees the efficacy and the accuracy of the predicted
function. Comparing with the Universal Approximation Theorem for
artificial neural network, whose proof takes considerable effort, our
theorem comes directly from Cauchy Integration formula (eq \ref{eq:int}).

In section \ref{sec:cauchy}, we start with Cauchy integral for complex
analytical functions, leading us to the derivation the Cauchy
Approximation Theorem. This theorem serves as the mathematical
fundation of our algorithm. Theoretically, our algorithm can achieve a
convergence rate of arbitrarily high order in any space dimensions.

In Sections \ref{sec:application}, we evaluate our algorithm
using a series of test cases. Observers are manually
positioned within the space, and some known functions are employed to
generate datasets in both one-dimensional and two-dimensional
spaces. The predicted functions are then compared with the actual
functions. As anticipated, the results are exceptional. We also
conducted tests on datasets containing random noises, and the outcomes
were quite satisfactory. The algorithm demonstrates impressive
predictive capabilities when it processes half-sided data to generate a
complete function.


We anticipate that this innovative algorithm will find extensive
applications in fields such as computational mathematics, machine
learning, and artificial intelligence. This paper provides a
fundamental principle for the algorithm.

\


We will formally state the fundamental theory behind our algorithm.
  
Given a function \( f \) holomorphic in a compact domain \( U \subset
\mathbb{C}^N \) in the complex $N$ dimensional space. For simplicity,
we assume that $U$ has a product structure, i.e., $U= U_1 \times U_2\times
\ldots \times U_n$, where each $U_k$, $k=1, \ldots, N$ is a compact domain in the
complex plane.  Let $P$ be the surface defined by

$$P = \partial U_1
\times \partial U_2 \times \ldots \times \partial U_N$$
Then a multi-dimensional Cauchy integral formula for $U$ is given by:
\begin{equation}\label{eq:int}
f(z_1, \ldots, z_N) = \left(\frac{1}{2\pi i}\right)^N \int \cdots \int_{P} \frac{f(\xi_1, \ldots, \xi_N)}{(\xi_1 - z_1) \cdots (\xi_N - z_N)} \, d\xi_1 \cdots d\xi_N,
\end{equation}
for all \( (z_1, \ldots, z_N) \in U \).

\

{\em The magic of the Cauchy Integration Formula lies in its ability to
determine the value of a function at any point using known values of
the function. This concept is remarkably similar to the principles of
machine learning!}

\
  
The Cauchy integral can be approximated, to any prescribed precision,  by a Riemann sum over a finite
number of points on $P$.  We can simplify the resulting Riemann sum in
the following form: 
\begin{equation} 
f(z_1, z_2, \ldots z_N)  \approx \sum_{k=1}^{m}\frac{\lambda_k}{(\xi_1^k -
  z_1)(\xi_2^k - z_2) \cdots (\xi_N^k -z_N)}, \label{eq:main}
\end{equation}
where 
$\lambda_1, \lambda_2, \ldots, \lambda_m$ are parameters depending on
the functional values at the sample points on $P$.

The Cauchy integral Formula guarantees the
accuracy of the above approximation
if enough points on $P$ are taken. However, there is no
reason at all that the sample points has to be on $P$ to achieve the
best approximation. Indeed, the surface $P$ itself is quite
arbitrary. 

We can state our fundamental theorem.

\begin{thm}[Cauchy Approximation Theorem] \label{thm:cauchy}
  \label{thm:CAT} Let $f(z_1, z_2, \ldots
  z_N)$ be an analytic function in an open domain $U \subset \C^N$ and
  let $M \subset U$ be a compact subset in $U$. 
  Given any $\ep >0$, there is a list of points $(\xi_1^k, \ldots,
  \xi_N^k)$, for $k=1, 2, \ldots, m$, in $U$ and corresponding
  parameters $\lambda_1, \lambda_2, \ldots, \lambda_m$, such that
\begin{equation} 
\vline \, f(z_1, z_2, \ldots z_N)  -\sum_{k=1}^{m}\frac{\lambda_k}{(\xi_1^k -
  z_1)(\xi_2^k - z_2) \cdots (\xi_N^k -z_N)} \,\vline < \ep, \label{eq:03}
\end{equation}
for all points $(z_1, z_2, \ldots z_N) \in M$.
\end{thm}

As we have explained, the proof is a simple application of the Cauchy
Theorem. We omit the details of the proof. We remark that, for
any given $\ep >0$, the number of points $m$ needed is approximately
at the level of $m \sim \ep^{-N}$, or equivalently, the error is
approximately at $\ep \sim m^{-1/N}$.  In fact, due to the nature of
complex analytic functions, where the size of the function also bounds the
derivative of the function, the error is much smaller. In fact, for
any fixed integer $k>0$, one can
show in theory that $\ep \sim o(m^{-k})$ for large $m$. \\

\

The Cauchy approximation is stated for complex analytical
functions. If the original function is real, it can be simplified as
follows.

\begin{equation}\label{eq:cauchy_general}
 f(x_1, x_2, \ldots x_N) \approx \mbox{Re} \left( \sum_{k=1}^{m}\frac{\lambda_k}{(\xi_1^k -
  x_1)(\xi_2^k - x_2) \cdots (\xi_N^k -x_N)} \right) , 
\end{equation}
where $x_1, x_2, \ldots, x_N$ are real, $\lambda_k, \; k=1, \ldots, m$
and $\xi_i^j, \; i=1, \ldots N, \;
j = 1, \ldots, m$ are complex numbers with positive imaginary
parts, Re($z$) is the real part of $z$.

\

A special simple case is when $N=1$, a function of one complex or
real variable. In this case, the approximation is

\begin{equation}\label{eq:cauchy_one_variable}
 f(x)  \approx \mbox{Re} \left( \sum_{k=1}^{m}\frac{\lambda_k}{\xi_k -
  x} \right), 
\end{equation}
with $x$ real and $\xi_k, \lambda_k,\; k = 1, \ldots m$ complex.

\

To derive the activation function, we express the complex parameters \(\xi_k\) and weights \(\lambda_k\) in terms of their real and imaginary parts:
\[
\xi_k = \xi_{k,\text{real}} + i \cdot \xi_{k,\text{imag}}, \quad \lambda_k = \lambda_{k,\text{real}} + i \cdot \lambda_{k,\text{imag}}.
\]

The denominator \(|\xi_k - x|^2\) can be written as:
\[
|\xi_k - x|^2 = (\xi_{k,\text{real}} - x)^2 + (\xi_{k,\text{imag}})^2.
\]

The fraction \(\frac{\lambda_k}{\xi_k - x}\) is expanded as:
\[
\frac{\lambda_k}{\xi_k - x} = \frac{\lambda_{k,\text{real}} + i \cdot \lambda_{k,\text{imag}}}{(\xi_{k,\text{real}} - x) + i \cdot \xi_{k,\text{imag}}}.
\]

Taking the real part, we obtain:
\[
\text{Re}\left(\frac{\lambda_k}{\xi_k - x}\right) = \frac{\lambda_{k,\text{real}} \cdot (\xi_{k,\text{real}} - x) + \lambda_{k,\text{imag}} \cdot \xi_{k,\text{imag}}}{(\xi_{k,\text{real}} - x)^2 + (\xi_{k,\text{imag}})^2}.
\]

To simplify further, assume that the real and imaginary parts of \(\lambda_k\) are trainable parameters, denoted as \(\lambda_1\) and \(\lambda_2\), and let \(\xi_{k,\text{imag}} = d\), where \(d\) is another trainable parameter. Substituting these into the expression, we obtain the activation function:
\[
\phi_{\lambda_1, \lambda_2, d}(x) = \frac{\lambda_1 \cdot x}{x^2 + d^2} + \frac{\lambda_2}{x^2 + d^2}.
\]

Explanation of Trainable Parameters:

\begin{enumerate}
\item \(\lambda_1\): This controls the linear contribution of the
  input \(x\). Every function can be decomposed into odd and even
  parts, this term captures the odd part.

\item \(\lambda_2\): This adjusts the constant term, providing
  additional flexibility in shaping the activation. This term captures
  the even part of the function.

\item \(d\): This parameter defines the scale of the denominator, controlling the range and smoothness of the activation function.

\end{enumerate}

\begin{figure}[h!]
    \centering
    \includegraphics[width=0.8\textwidth]{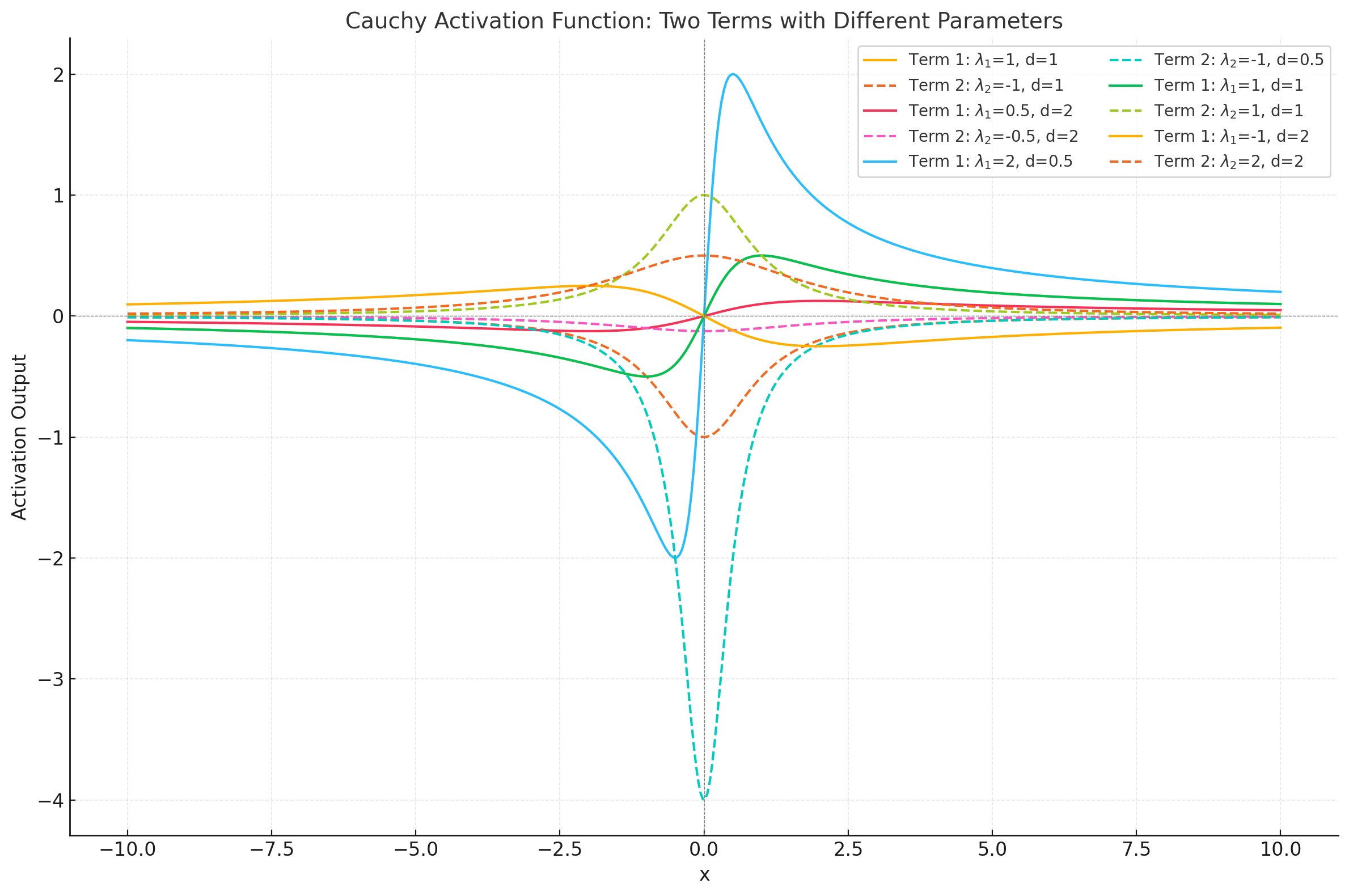}
    \caption{Visualization of the two terms of the Cauchy activation function under different parameter settings.}
    \label{fig:cauchy_activation}
\end{figure}

The Cauchy activation function provides unique advantages due to its flexibility and localization properties, making it particularly suitable for tasks involving smooth or highly localized data patterns. For effective initialization, we typically set \(\lambda_1\) and \(\lambda_2\) to small positive values (e.g., around 0.01), which ensures minimal bias and avoids large gradients during early training. The parameter \(d\) is initialized to 1, providing a balance between smoothness and localization. These initial settings allow the activation function to stabilize training dynamics and adaptively learn optimized parameters to enhance model performance.



\

Cauchy approximation theorem can be easily implemented into an
artificial neural network (\cite{ZLX24}). The resulting
network, termed CauchyNet, is very efficient for lower dimensional
problems, say for $N \leq 10$. However, for large $m$, the
multiplicative terms in the denominator pose serious computational
difficulties. As typical computer vision problems are high
dimensional, with the input data for a $30 \times 30$ pixel
image being at least 900 dimensional, we need a different algorithm to
handle high dimensional problems.

Towards this end, we prove a general approximation theorem. The
central idea is that, if one can approximate one-dimensional function
with certain functional class, then one can extend the functional class to
approximate functions in any dimension. Mathematically, this
corresponds to approximation in dual spaces. This method is
particularly effective for feature-capturing.

As an example, Cauchy approximation can approximate any one
dimensional function, as we see in the above theorem, therefore it can
approximate any dimensioanl function through linear combinations.

More precisely, we consider $C(\R)= C(\R, \R)$, the set of continuous
real-valued functions in $\R$. A family of real valued functions on
$\R$, $\Phi = \{\phi_a \; | \; a \in A\}$, where $A$ is some index
set, is said to have {\em universal approximation property}\/ or
simply {\em approximation property}\/, if for any closed bounded
interval $I$ in $\R$ and any continuous function $g \in C(I, \R)$,
there is a sequence of functions
$g^j = a_1^j g_{a_1} + a_2^j g_{a_2} + \dots + a_{k_j}^j g_{a_{k_j}}$
such that $g^j \ra g$ uniformly over $I$. In other words, $\Phi$ has
approximation property if every continuous function can be
approximated by a linear combination of functions in $\Phi$, uniformly
over a bounded interval.

The main result is the following general approximation theorem.

\begin{thm}[General  Approximation Theorem] \label{thm:main}
  Let $\Phi$ be a family of functions in \textcolor{black}{$C(\R, \R)$} with the \textcolor{black}{universal approximation
  property}. Let
  $$\Phi^N =  \{ \phi_a(a_1x_1 + a_2x_2 + \ldots + a_Nx_N) \; | \;
  (a_1, \ldots, a_N) \in \R^N, \phi_a \in \Phi\}$$ Then, the family of
  functions $\Phi^N$ has the \textcolor{black} {universal approximation property} in \textcolor{black}{$C(\R^N, \R)$}, i.e,
  every continuous function in $\R^N$ can be approximated by a linear
  combination of functions in $\Phi^N$, uniformly over a compact
  subset in $\R^N$.
\end{thm}

The proof of this theorem can be found in Appendix A.

The Cauchy Approximation Theorem and the General Approximation Theorem
demonstrate that the Cauchy activation function can be used
effectively, at least mathematically, to approximate functions of any
dimension. We will test its practical applications in the following
sections.

Compared to popular neural networks, it appears that only the
activation function needs to be changed. However, due to the novelty
of the Cauchy activation function, structural changes to the neural
networks are necessary to leverage its intrinsic
efficiencies. Generally, we can significantly simplify complex
networks to achieve comparable or better results. This theoretical and
operational framework prompts us to rename the network to 'XNet' to
reflect the fundamental improvements in our networks capabilities.

\section{Examples} \label{sec:application}

In this section, we focus on image classification and PDEs. Regression tasks, especially function fitting and computations for other PDEs, are also highly relevant and can be explored further, as detailed in \cite{Li2024}. These tasks serve as important benchmarks for evaluating the approximation capabilities of neural networks and their ability to generalize under different problem settings.

Our approach was primarily compared with MLP,  where it demonstrated significant advantages in performance. Specifically, the Cauchy activation function showcased superior approximation power in regression tasks by achieving high-order accuracy in function fitting. This is particularly evident in its ability to handle complex target functions with minimal error, as well as its robustness in noisy data scenarios.

Moreover, when applied to PDEs, the method consistently achieved
higher accuracy, usually much more than 10 times while achieved in much
less computing time, in solving both linear and nonlinear equations, underscoring its potential in scientific computing applications. These results demonstrate the versatility of our approach across diverse tasks, and future work will further investigate its applicability in more challenging domains.


It is important to note that Theorem \ref{thm:main} applies to any higher orders. When solving PDEs, the higher-order accuracy of the solution is a key criterion for evaluating the method. As we optimize through parameter tuning, we will observe increasingly higher-order approximation effects. With the amount of data we have, we have already seen high-order accuracy. Since our theorem applies to arbitrary higher orders, we can observe even higher-order accuracy through parameter tuning and other techniques. This will be a focus of our future work.

\subsection{Regression Task}\label{subsec:regression}

Our approach utilizes the Cauchy activation function known for its ability to achieve high-order accuracy in function fitting. 
\subsubsection{High-Order Approximation Analysis}

In a noise-free environment, we define the target function for the regression task as:
\[
y_{\text{train}} = x_1^2 - x_1 \cdot x_2 + 3 \cdot x_2 + x_2^2 + \frac{1}{5 + x_1^2}.
\]
We conducted training with a dataset size of \(N = 2500\), using a single hidden layer with 400 neurons, over 12000 epochs, and a learning rate of 0.001. The observed MSE was remarkably low at \(1.7 \times 10^{-6}\), indicating a high level of model accuracy.

The error dynamics for the Cauchy model are given by:
\[
\text{MSE} = O\left(\frac{1}{N} + \frac{1}{h^p}\right),
\]
where \( h = 400 \) is the number of neurons, and \( p \) represents the order of approximation. Solving for \( p \), we find:
\[
1.7 \times 10^{-6} = \frac{1}{2500} + \frac{1}{400^p},
\]
resulting in \( p \approx 12.8 \), a clear demonstration of high-order approximation capability.

\subsubsection{Comparison with ReLU under Noisy Conditions}

To simulate real-world scenarios in our regression tasks, we introduced Gaussian noise to the target function:
\[
y_{\text{train}} = x_1^2 - x_1 \cdot x_2 + 3 \cdot x_2 + x_2^2 + \frac{1}{5 + x_1^2} + \mathcal{N}(0, \sigma^2),
\]
where \(\mathcal{N}(0, \sigma^2)\) denotes Gaussian noise with a standard deviation of \(\sigma = 0.1\), representing a low noise scenario.

In a comparative study over 1000 epochs, we assessed the performance of neural networks equipped with Cauchy and ReLU activation functions.  We used a fixed learning rate of \(lr = 0.001\) across all experiments. Although we tested various learning rates, \(lr = 0.001\) was chosen as it allowed us to clearly discern performance differences between the activation functions within 1000 epochs.

\begin{table}[ht]
\centering
\caption{Loss Comparison: Cauchy vs. ReLU}
\begin{tabular}{|c|c|c|c|}
\hline 
\textbf{Activation} & \textbf{Hidden Dim} & \textbf{Noise Level} & \textbf{Loss} \\ 
\hline 
Cauchy & 400 & Clean & 0.000538 \\ 
\hline
Cauchy & 400 & 10\% Noise & 0.010413 \\ 
\hline
ReLU & 400 & Clean & 0.010158 \\ 
\hline
ReLU & 400 & 10\% Noise & 0.021373 \\ 
\hline
Cauchy & 800 & Clean & 0.000258 \\ 
\hline
Cauchy & 800 & 10\% Noise & 0.010155 \\ 
\hline
ReLU & 800 & Clean & 0.001646 \\ 
\hline
ReLU & 800 & 10\% Noise & 0.012083 \\ 
\hline
\end{tabular}
\end{table}

As we noted eralier, if we run more epoches with smaller learning
rate, MSE with Cauchy activation can be further reduced to the order
of $10^{-6}$, clearly showing the high order effect of Cauchy approximation. The data clearly demonstrate the superior performance of the Cauchy activation function, which maintained lower loss values across both noise levels and hidden dimensions, confirming its effectiveness and robustness in noisy conditions.

\subsection{ Handwriting Recognition: MNIST with XNet}

The XNet architecture used for these experiments consists of an input layer \(W_{in}\), a hidden layer \(W\), and an output layer \(W_{out}\). The input \(28 \times 28\) grayscale image is reshaped into a \(784 \times 1\) vector, which is compressed by the input layer into \(N\) features and processed by the \(N \times N\) hidden layer. The output layer maps these \(N\) features to 10 classes representing digits 0--9. The model was implemented in PyTorch and trained using the ADAM optimizer with a learning rate of 0.0001.

We evaluated XNet on the MNIST dataset using two fully connected architectures--a single-layer network ([100]) and a two-layer network ([128, 64])--as well as a CNN with a simple multi-scale kernel size convolution layer. Across all architectures, the Cauchy activation function demonstrated superior performance in accuracy, F1 score, AUC, and other metrics on the testing data. These results underscore that the Cauchy activation outperforms other activation functions across multiple benchmarks, including accuracy, loss minimization, convergence speed, generalization error, F1 score, and AUC, albeit with a slightly longer runtime. The slightly longer runtime is due to the custom implementation of the Cauchy activation function compared to the optimized PyTorch-packaged activations.

In the [100] network, the Cauchy activation achieved a validation accuracy of \(96.33\%\), a validation loss of \(0.1379\), an F1 score of \(0.964\), and an AUC of \(0.987\), significantly outperforming ReLU (\(95.27\%\), \(0.1693\), \(0.952\), \(0.978\)) and Sigmoid (\(90.73\%\), \(0.3289\), \(0.907\), \(0.940\)). Similarly, in the [128, 64] network, Cauchy led with a validation accuracy of \(96.87\%\), a validation loss of \(0.1434\), an F1 score of \(0.969\), and an AUC of \(0.990\), surpassing ReLU (\(96.22\%\), \(0.1628\), \(0.962\), \(0.985\)) and Leaky ReLU (\(96.46\%\), \(0.1547\), \(0.965\), \(0.987\)). While the Cauchy activation incurred a slightly longer runtime due to our custom implementation compared to PyTorch's built-in functions, its substantial improvements in accuracy, generalization, and other metrics validate its efficiency and robustness for diverse applications.

\begin{figure}[H]
    \centering
 
    \begin{minipage}[b]{0.48\textwidth}
        \centering
        \includegraphics[width=\linewidth]{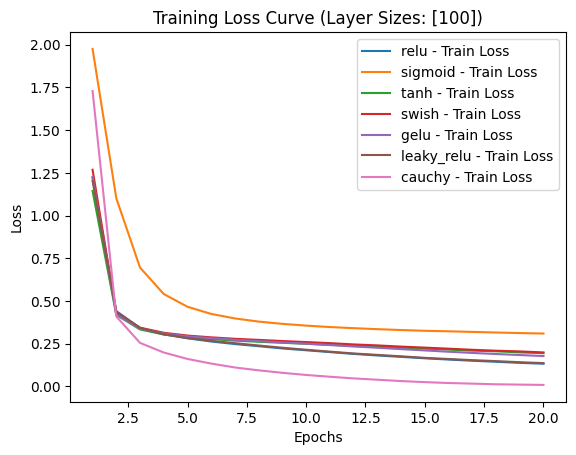}
        \caption{Training Loss for [100].}
        \label{fig:train_loss_100}
    \end{minipage}\hfill
    \begin{minipage}[b]{0.48\textwidth}
        \centering
        \includegraphics[width=\linewidth]{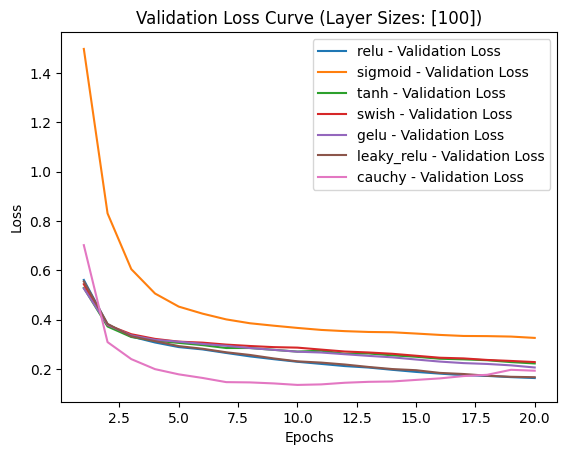}
        \caption{Validation Loss for [100].}
        \label{fig:valid_loss_100}
    \end{minipage}
    \vspace{0.5cm}
    
    \begin{minipage}[b]{0.48\textwidth}
        \centering
        \includegraphics[width=\linewidth]{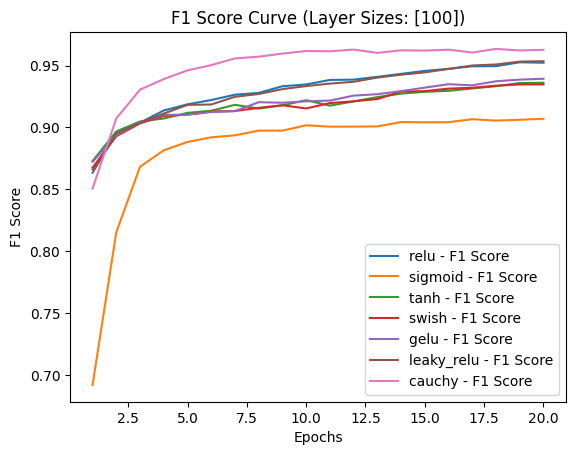}
        \caption{F1 Score for [100].}
        \label{fig:f1_100}
    \end{minipage}\hfill
    \begin{minipage}[b]{0.48\textwidth}
        \centering
        \includegraphics[width=\linewidth]{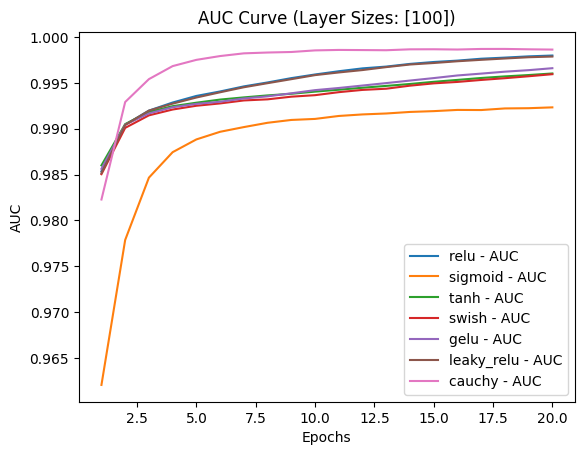}
        \caption{AUC for [100].}
        \label{fig:auc_100}
    \end{minipage}
    \caption{Performance Metrics for the [100] Model.}
    \label{fig:metrics_100}
\end{figure}

\begin{figure}[H]
    \centering
    
    \begin{minipage}[b]{0.48\textwidth}
        \centering
        \includegraphics[width=\linewidth]{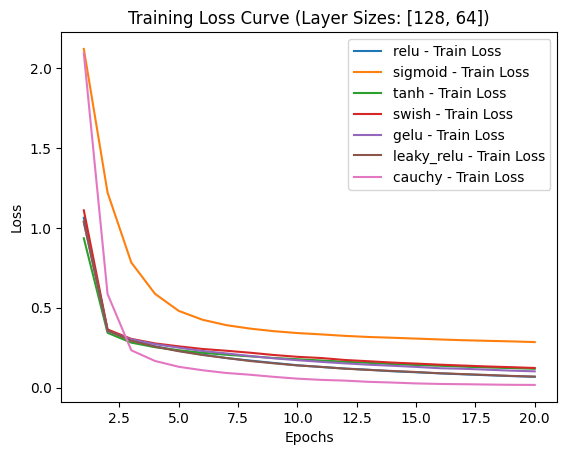}
        \caption{Training Loss for [100].}
        \label{fig:train_loss_2layer.}
    \end{minipage}\hfill
    \begin{minipage}[b]{0.48\textwidth}
        \centering
        \includegraphics[width=\linewidth]{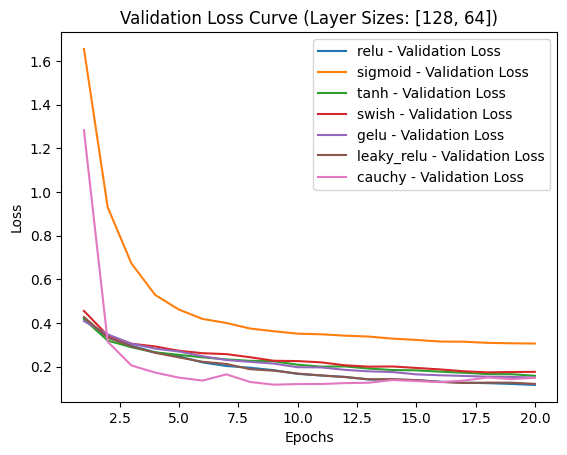}
        \caption{Validation Loss for [128,64].}
        \label{fig:valid_loss_2layer.}
    \end{minipage}
    \vspace{0.5cm}
    
    \begin{minipage}[b]{0.48\textwidth}
        \centering
        \includegraphics[width=\linewidth]{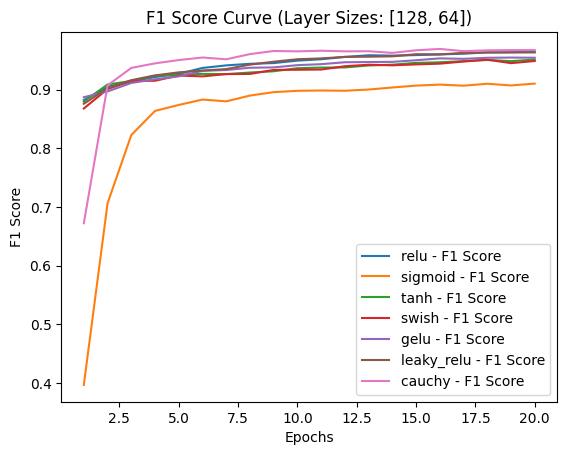}
        \caption{F1 Score for [128,64].}
        \label{fig:f1_2layer}
    \end{minipage}\hfill
    \begin{minipage}[b]{0.48\textwidth}
        \centering
        \includegraphics[width=\linewidth]{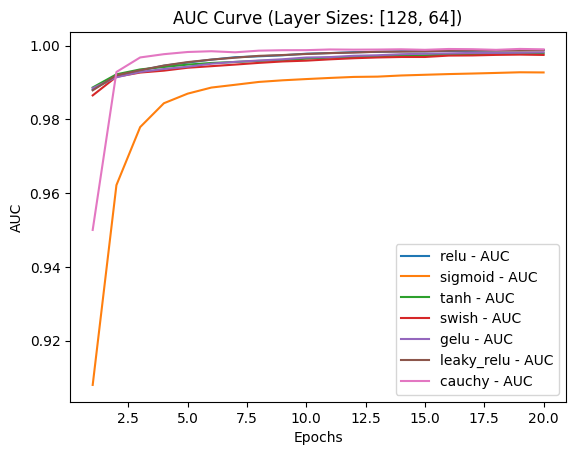}
        \caption{AUC for [128,64].}
        \label{fig:auc_2layer.}
    \end{minipage}
    \caption{Performance Metrics for the [128,64] Model.}
    \label{fig:metrics_2layer.}
\end{figure}

Figures \ref{fig:train_loss_100}, \ref{fig:train_loss_2layer.}, \ref{fig:valid_loss_100}, and \ref{fig:valid_loss_2layer.} depict the training and validation loss curves with learning rate 0.001 (results for learning rate 0.01 show similar patterns and are omitted for brevity). The Cauchy activation (pink curve) demonstrates notably faster convergence during training, not only achieving near-zero training loss more quickly but also reaching optimal validation metrics in earlier epochs. This rapid convergence to optimal performance is particularly valuable in practical applications. Sigmoid (orange curve), on the other hand, converges slowly and suffers from high training and validation losses due to vanishing gradients. While ReLU and Leaky ReLU show competitive final performance, Cauchy outperforms them in terms of convergence speed and the ability to achieve better metrics earlier in the training process.

Cauchy's efficiency is further highlighted by its ability to achieve high validation accuracy, low validation losses, and superior F1 and AUC scores in earlier epochs compared to other activation functions. This characteristic makes it particularly well-suited for tasks requiring both high precision and efficient training, such as medical diagnosis or fraud detection. While the final epoch results show some signs of overfitting, the superior performance achieved in earlier epochs demonstrates Cauchy's potential to reach optimal solutions more efficiently. Although Leaky ReLU achieves competitive validation loss in later epochs, it requires more training time to reach comparable performance levels. Sigmoid consistently exhibits the weakest generalization throughout the training process.

These results underscore the potential of the Cauchy activation function not only in achieving superior metrics but also in reaching optimal performance more efficiently, making it a promising choice for neural network architectures, particularly in scenarios requiring both high precision and training efficiency. The rapid convergence to optimal performance suggests that with proper regularization and early stopping strategies, Cauchy activation could provide both better results and more efficient training processes.

\begin{table}[H]
\centering
\scriptsize
\begin{tabular}{|c|c|c|c|c|c|c|c|c|c|}
\hline
\textbf{Params} & \textbf{Epoch} & \textbf{Train Loss} & \textbf{Train Acc} & \textbf{Val Loss} & \textbf{Val Acc} & \textbf{F1 Score} & \textbf{AUC} & \textbf{Gen Error} & \textbf{Time (s)} \\ \hline
$[100], 0.001, \text{relu}$    & 20 & 0.1337 & 0.9643 & 0.1651 & 0.9524 & 0.9522 & 0.9979 & 0.0120 & 1.08 \\ \hline
$[100], 0.001, \text{sigmoid}$ & 20 & 0.3088 & 0.9165 & 0.3259 & 0.9079 & 0.9070 & 0.9923 & 0.0086 & 1.21 \\ \hline
$[100], 0.001, \text{tanh}$    & 20 & 0.1938 & 0.9453 & 0.2224 & 0.9363 & 0.9361 & 0.9960 & 0.0090 & 1.09 \\ \hline
$[100], 0.001, \text{swish}$    & 20 & 0.1984 & 0.9444 & 0.2279 & 0.9351 & 0.9348 & 0.9959 & 0.0093 & 1.23 \\ \hline
$[100], 0.001, \text{gelu}$     & 20 & 0.1770 & 0.9513 & 0.2060 & 0.9397 & 0.9393 & 0.9966 & 0.0116 & 1.10 \\ \hline
$[100], 0.001, \text{leaky\_relu}$ & 20 & 0.1356 & 0.9645 & 0.1666 & 0.9538 & 0.9537 & 0.9979 & 0.0107 & 1.07 \\ \hline
$[100], 0.001, \text{cauchy}$   & 20 & 0.0079 & 0.9982 & 0.1929 & 0.9630 & 0.9627 & 0.9986 & 0.0352 & 1.20 \\ \hline
$[100], 0.01, \text{relu}$    & 20 & 0.0755 & 0.9785 & 0.1355 & 0.9594 & 0.9592 & 0.9989 & 0.0192 & 1.09 \\ \hline
$[100], 0.01, \text{sigmoid}$ & 20 & 0.2137 & 0.9394 & 0.2382 & 0.9313 & 0.9308 & 0.9962 & 0.0082 & 1.15 \\ \hline
$[100], 0.01, \text{tanh}$    & 20 & 0.0989 & 0.9705 & 0.1296 & 0.9606 & 0.9606 & 0.9988 & 0.0098 & 1.08 \\ \hline
\end{tabular}
\caption{Performance Comparison of Activation Functions (Part 1: $[100], 0.001$ and $[100], 0.01$)}
\label{tab:activations_part1}
\end{table}
\begin{table}[H]
\centering
\scriptsize
\begin{tabular}{|c|c|c|c|c|c|c|c|c|c|}
\hline
\textbf{Params} & \textbf{Epoch} & \textbf{Train Loss} & \textbf{Train Acc} & \textbf{Val Loss} & \textbf{Val Acc} & \textbf{F1 Score} & \textbf{AUC} & \textbf{Gen Error} & \textbf{Time (s)} \\ \hline
$[128, 64], 0.001, \text{relu}$    & 20 & 0.0687 & 0.9811 & 0.1165 & 0.9663 & 0.9663 & 0.9989 & 0.0147 & 1.15 \\ \hline
$[128, 64], 0.001, \text{sigmoid}$ & 20 & 0.2849 & 0.9183 & 0.3011 & 0.9117 & 0.9106 & 0.9929 & 0.0066 & 1.15 \\ \hline
$[128, 64], 0.001, \text{tanh}$    & 20 & 0.1169 & 0.9666 & 0.1603 & 0.9532 & 0.9531 & 0.9978 & 0.0135 & 1.14 \\ \hline
$[128, 64], 0.001, \text{swish}$    & 20 & 0.1217 & 0.9652 & 0.1663 & 0.9513 & 0.9512 & 0.9978 & 0.0140 & 1.11 \\ \hline
$[128, 64], 0.001, \text{gelu}$     & 20 & 0.0990 & 0.9713 & 0.1454 & 0.9548 & 0.9546 & 0.9983 & 0.0166 & 1.11 \\ \hline
$[128, 64], 0.001, \text{leaky\_relu}$ & 20 & 0.0536 & 0.9826 & 0.1354 & 0.9606 & 0.9601 & 0.9988 & 0.0220 & 1.27 \\ \hline
$[128, 64], 0.001, \text{cauchy}$   & 20 & 0.0632 & 0.9796 & 0.1141 & 0.9671 & 0.9671 & 0.9992 & 0.0125 & 1.35 \\ \hline
$[128, 64], 0.01, \text{relu}$    & 20 & 0.0503 & 0.9837 & 0.1066 & 0.9660 & 0.9656 & 0.9993 & 0.0176 & 1.12 \\ \hline
$[128, 64], 0.01, \text{sigmoid}$ & 20 & 0.1667 & 0.9499 & 0.1826 & 0.9460 & 0.9458 & 0.9978 & 0.0038 & 1.15 \\ \hline
$[128, 64], 0.01, \text{tanh}$    & 20 & 0.0682 & 0.9785 & 0.1207 & 0.9627 & 0.9622 & 0.9989 & 0.0158 & 1.14 \\ \hline
$[128, 64], 0.01, \text{leaky\_relu}$ & 20 & 0.0536 & 0.9839 & 0.1162 & 0.9667 & 0.9666 & 0.9989 & 0.0173 & 1.12 \\ \hline
$[128, 64], 0.01, \text{swish}$    & 20 & 0.0650 & 0.9785 & 0.1157 & 0.9654 & 0.9653 & 0.9991 & 0.0131 & 1.25 \\ \hline
$[128, 64], 0.01, \text{gelu}$     & 20 & 0.0521 & 0.9824 & 0.0993 & 0.9678 & 0.9676 & 0.9993 & 0.0147 & 1.13 \\ \hline
$[128, 64], 0.01, \text{cauchy}$   & 20 & 0.0632 & 0.9796 & 0.1141 & 0.9671 & 0.9671 & 0.9992 & 0.0125 & 1.35 \\ \hline
\end{tabular}
\caption{Performance Comparison of Activation Functions (Part 2: $[128, 64], 0.001$ and $[128, 64], 0.01$)}
\label{tab:activations_part2}
\end{table}

As shown in Tables 1 and 2, the Cauchy activation function demonstrates superior performance across different network architectures and learning rates. With the [100] architecture and learning rate of 0.001, it achieves remarkable training accuracy of $99.82\%$ and validation accuracy of $96.30\%$, significantly outperforming other activation functions. Similarly impressive results are observed with the [128, 64] architecture, where Cauchy maintains consistently high performance with $97.96\%$ training accuracy and $96.71\%$ validation accuracy at learning rate 0.01. The higher generalization error observed (e.g., 0.0352 for [100] architecture) can be attributed to the rapid convergence and the absence of regularization techniques in our comparative setup, rather than an inherent limitation of the activation function. Notably, Cauchy activation achieves superior AUC scores (0.9986-0.9992) across all configurations, indicating excellent classification reliability. While the computational time appears longer in our experiments, this is primarily due to our custom implementation of the Cauchy activation function compared to PyTorch's highly optimized built-in functions for other activations, rather than an inherent computational complexity of the function itself. A native implementation would likely achieve comparable computational efficiency. These results suggest that the Cauchy activation function, when properly regularized and efficiently implemented, could be the optimal choice among common activation functions for similar classification tasks.

In the 2nd experiment, we used a simple convolutional neural network (CNN) with a single convolutional layer, followed by a max-pooling layer and a fully connected layer.

\begin{figure}[h!]
    \centering
    \includegraphics[width=0.8\textwidth]{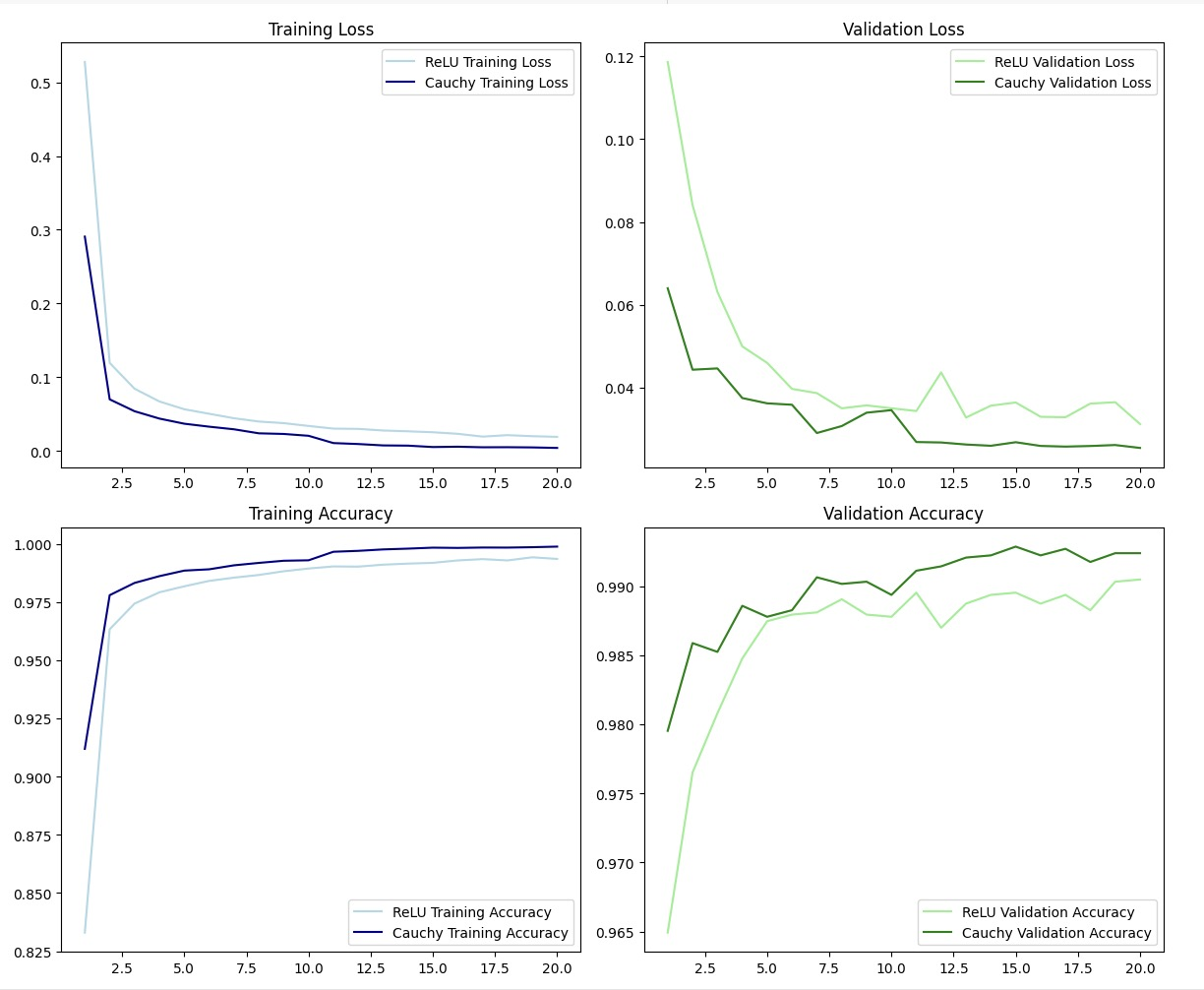}
    \caption{Training and validation performance of CNN with ReLU and Cauchy activation functions.}
    \label{fig:performance}
\end{figure}

We experimented with two different activation functions: ReLU and Cauchy. In our model configuration, we replaced the ReLU activation function with the Cauchy activation function after the third convolutional layer. The learning rate was set to 0.001 for both models. The training and validation losses, as well as accuracies after 20 epochs, are presented in Table~\ref{table:results}.

\begin{table}[htbp]
    \centering
    \caption{Performance comparison of CNN with ReLU and Cauchy activation functions after 20 epochs, using a 3-layer convolutional architecture.}
    \label{table:results}
    \begin{tabular}{|c|c|c|c|c|c|}
        \hline
        \textbf{Activation} & \textbf{LR} & \textbf{Training Loss} & \textbf{Training Acc} & \textbf{Valid Loss} & \textbf{Valid Acc} \\
        \hline
        ReLU & 0.001 &  0.0190 & 0.9935 & 0.0312 & 0.9905 \\
        \hline
        Cauchy & 0.001 & 0.0039 & 0.9988 & 0.0254 & 0.9924 \\
        \hline
    \end{tabular}
\end{table}

As shown in Table~\ref{table:results}, the model using the Cauchy activation function achieved a training loss of 0.0039 and a training accuracy of $99.88\%$, while the validation loss was 0.0254 and the validation accuracy was $99.24\%$. On the other hand, the model using the ReLU activation function resulted in a training loss of 0.0190 and a training accuracy of $99.35\%$, with a validation loss of 0.0312 and a validation accuracy of $99.05\%$.

These results suggest that the Cauchy activation model outperformed the ReLU model in both training and validation, showing lower losses and higher accuracies, indicating better generalization capabilities.

\textbf{Remark}

The fluctuations observed in the validation loss curve for the Cauchy activation function are a result of its unique derivative properties. Unlike ReLU, whose derivative is piecewise constant, the derivative of the Cauchy activation function exhibits significant variations across the input space. This characteristic allows the Cauchy activation function to capture complex, nonlinear relationships more effectively, particularly in challenging regions of the loss landscape.  

These fluctuations can be more pronounced when using a relatively larger learning rate, as the optimizer takes larger steps during training, amplifying the sensitivity to the Cauchy activation's gradient dynamics. Importantly, reducing the learning rate mitigates this behavior, leading to smoother convergence. The trade-off lies in the balance between learning speed and stability: smaller learning rates improve stability, while larger rates accelerate convergence but may introduce short-term oscillations.  

Despite these initial fluctuations, the validation accuracy curve shows stable improvement, indicating that the Cauchy activation function's high-order approximation capability ultimately enables superior performance. 

\subsection{CIFAR-10}

We conducted a comprehensive study to evaluate the performance of the proposed Cauchy activation function compared to six widely used activation functions: ReLU, Sigmoid, Tanh, Swish, GeLU, and Leaky ReLU. The experiments were performed on the CIFAR-10 dataset, which contains 60,000 32x32 color images across 10 classes, making it a standard benchmark for image classification tasks.

\textbf{Experimental Setup:} 

\begin{itemize} \item \textbf{CNN Architecture:} A custom Convolutional Neural Network (CNN) with 6 convolutional layers followed by fully connected layers. For experiments with the proposed Cauchy activation function, the original three fully connected layers were replaced with a single fully connected layer. Additionally, a normalization block (NB) was introduced to enhance stability. The Cauchy activation was applied exclusively in this modified fully connected layer, leveraging its high-order approximation capability. For all other activation functions (ReLU, Sigmoid, Tanh, Swish, GeLU, Leaky ReLU), the architecture remained in its original form, with the activation functions directly replacing ReLU without altering the structure.

\item \textbf{ResNet9 Architecture:} A compact version of the ResNet architecture, designed for efficiency on smaller datasets. For experiments with the Cauchy activation function, the latter half of the convolutional layers and the residual connections were modified to use Cauchy activation. This adjustment aimed to explore the activation's impact on deeper network components. For all other activation functions, ReLU was directly replaced with the corresponding activation function throughout the network, maintaining the original structure and residual connections.

\end{itemize}

\textbf{Training Procedure:}
For both architectures, training was conducted in two phases:
\begin{enumerate}
    \item An initial training phase with \( 20 \) epochs using a learning rate of \( 0.001 \).
    \item A fine-tuning phase with \( 10 \) epochs using a reduced learning rate of \( 0.0001 \).
\end{enumerate}

\textbf{Results and Analysis:}

\begin{table}[h]
\centering
\caption{Comparison of Activation Functions on CIFAR-10 (6 Convolutional Layers, 30 Epochs)}
\label{tab:activation_comparison_cnn}
\begin{tabular}{|l|c|}
\hline
\textbf{Activation Function} & \textbf{Final Validation Accuracy (\%)} \\ \hline
ReLU                         & 78.60                                  \\ \hline
Sigmoid                      & 76.71                                   \\ \hline
Tanh                         & 76.74                                  \\ \hline
Swish                        & 78.54                                  \\ \hline
GeLU                         & 78.81                                  \\ \hline
Leaky ReLU                   & 79.29                                  \\ \hline
Cauchy                       & 81.90                                  \\ \hline
\end{tabular}
\end{table}

As shown in Table~\ref{tab:activation_comparison_cnn}, the Cauchy activation function achieves the highest accuracy of \textbf{81.90\%}, outperforming other activation functions such as ReLU (\textbf{78.60\%}) and GeLU (\textbf{78.81\%}). 

\begin{table}[h]
\centering
\caption{Performance Comparison of Activation Functions on CIFAR-10 with ResNet9}
\label{tab:activation_comparison_resnet9}
\begin{tabular}{|l|c|c|}
\hline
\textbf{Activation Function} & \textbf{Learning Rate} & \textbf{Validation Accuracy (\%)} \\ \hline
ReLU                         & 0.01                   & 90.91                              \\ \hline
ReLU                         & 0.005                  & 90.72                              \\ \hline
Sigmoid                      & 0.01                   & 80.30                              \\ \hline
Sigmoid                      & 0.005                  & 80.86                              \\ \hline
Tanh                         & 0.01                   & 85.99                              \\ \hline
Tanh                         & 0.005                  & 84.84                              \\ \hline
Swish                        & 0.01                   & 90.34                              \\ \hline
Swish                        & 0.005                  & 90.62                              \\ \hline
GeLU                         & 0.01                   & 90.57                              \\ \hline
GeLU                         & 0.005                  & 91.09                              \\ \hline
Leaky ReLU                   & 0.01                   & 90.56                              \\ \hline
Leaky ReLU                   & 0.005                  & 90.65                              \\ \hline
Cauchy                       & 0.01                   & 90.28                              \\ \hline
Cauchy                       & 0.005                  & \textbf{91.42}                     \\ \hline
\end{tabular}
\end{table}

In the ResNet9 experiments, summarized in Table~\ref{tab:activation_comparison_resnet9}, the Cauchy activation function achieves the highest accuracy (\textbf{91.42\%}) at \( \text{lr} = 0.005 \), outperforming GeLU (\textbf{91.09\%}). Lower learning rates consistently improve performance across all activation functions, highlighting the benefits of stable convergence.

\textbf{Conclusion:}
The results demonstrate the superiority of the Cauchy activation function in terms of accuracy and generalization. Its ability to achieve higher accuracy with fewer fully connected layers in CNN experiments further validates its theoretical high-order approximation capabilities. Moreover, its competitive performance in ResNet9 experiments suggests that Cauchy activation can be a promising alternative to standard activation functions in deep learning.

\subsection{PDE:  Heat Function}

The 1-dimensional heat equation in this example is given by:

\begin{equation}
    \frac{\partial u}{\partial x} - 2 \frac{\partial u}{\partial t} - u = 0,
    \label{eq:heat_equation}
\end{equation}

where \(u(x,t)\) is the temperature distribution function, \(x\) is the spatial coordinate, and \(t\) is the time.

The boundary and initial conditions are specified as follows:

\begin{itemize}
    \item Initial condition:
    \begin{equation}
        u(x, 0) = 6 e^{-3x}, \quad \text{for} \; 0 \le x \le 2.
    \end{equation}
    
    \item Boundary conditions:
    \begin{equation}
        u(0, t) = 0, \quad \text{for} \; 0 \le t \le 1,
    \end{equation}
    \begin{equation}
        u(2, t) = 0, \quad \text{for} \; 0 \le t \le 1.
    \end{equation}
\end{itemize}


We trained the original PINN with the sigmoid activation function and then modified the network to use a Cauchy activation function instead of the sigmoid function. The training process was repeated with the new activation function to compare the performance.

\begin{figure}[h!]
    \centering
    \begin{minipage}[t]{0.48\textwidth}
        \centering
        \includegraphics[width=\linewidth,height=4cm,keepaspectratio]{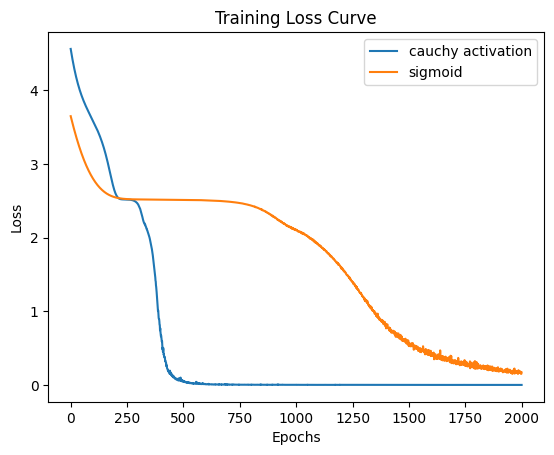} 
        \caption{loss curve by XNet and PINN}
        \label{fig:fem}
    \end{minipage}\hfill
    \end{figure}

\begin{table}[h]
\centering
\caption{Comparison with  PINN}
\begin{tabular}{|l|l|l|l|}
\hline
\textbf{Activation } & \textbf{Network Layers} & \textbf{Training Loss} & \textbf{Mean Error } \\ \hline
Sigmoid                     & 5 Layers FNN & 0.0064                 & 2e-3                                       \\ \hline
Cauchy                      & 5 Layers & 0.0003                 & 6e-5                                       \\ \hline
\end{tabular}
\end{table}

\subsection{Poisson Equation with Dirichelet Boundary Condition}

We study the Poisson equation 
\begin{equation}
\nabla^2 u(x, y) = f(x, y), \quad \text{for } (x, y) \in \Omega,
\end{equation}
with Dirichlet boundary conditions:
\begin{equation}
u(x, y) = 0, \quad \text{for } (x, y) \in \partial\Omega,
\end{equation}

where the source term \( f(x, y) = -8\pi^2 \sin(2\pi x) \sin(2\pi y) \).  The ground truth solution is $u(x,y)={\rm sin}(2\pi x){\rm sin}(2\pi y)$.

The dataset size is 2000, 1000 interior points and 1000 boundary points.

In this simple low-dimensional equation, we firstly found that using the least-squares method in MATLAB effectively solves the problem. We worked with 1000 interior points and 1000 boundary points, placing 400 observation points, which correspond to the boundary points in the Cauchy integral formula in complex space. The method is equivalent with our CauchyNet in \cite{ZLX24}. We defined the boundary as an ellipse. (If we optimize the boundary points, we can achieve even higher computational accuracy.) The results are shown below:

\begin{figure}[h!]
    \centering
    \begin{minipage}[t]{0.48\textwidth}
        \centering
        \includegraphics[width=\linewidth,height=4cm,keepaspectratio]{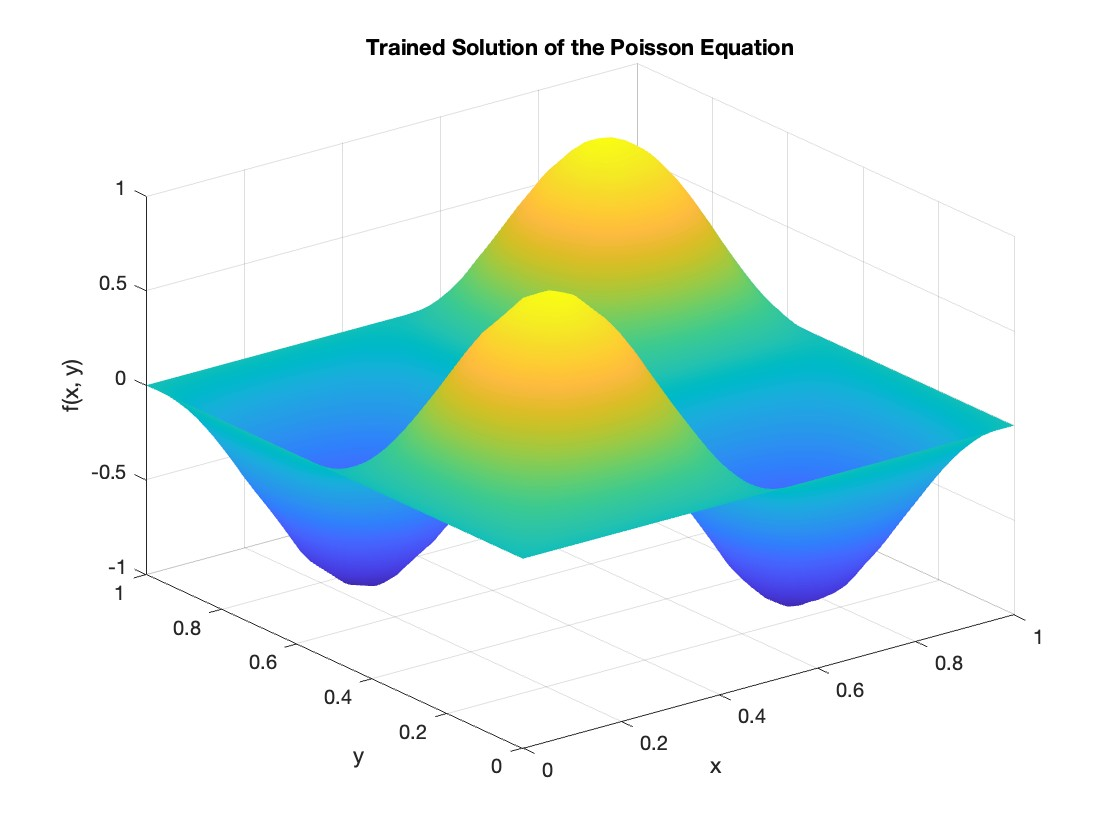}
        \caption{Cauchy activation}
        \label{fig:fem}
    \end{minipage}\hfill
    \begin{minipage}[t]{0.48\textwidth}
        \centering
        \includegraphics[width=\linewidth,height=4cm,keepaspectratio]{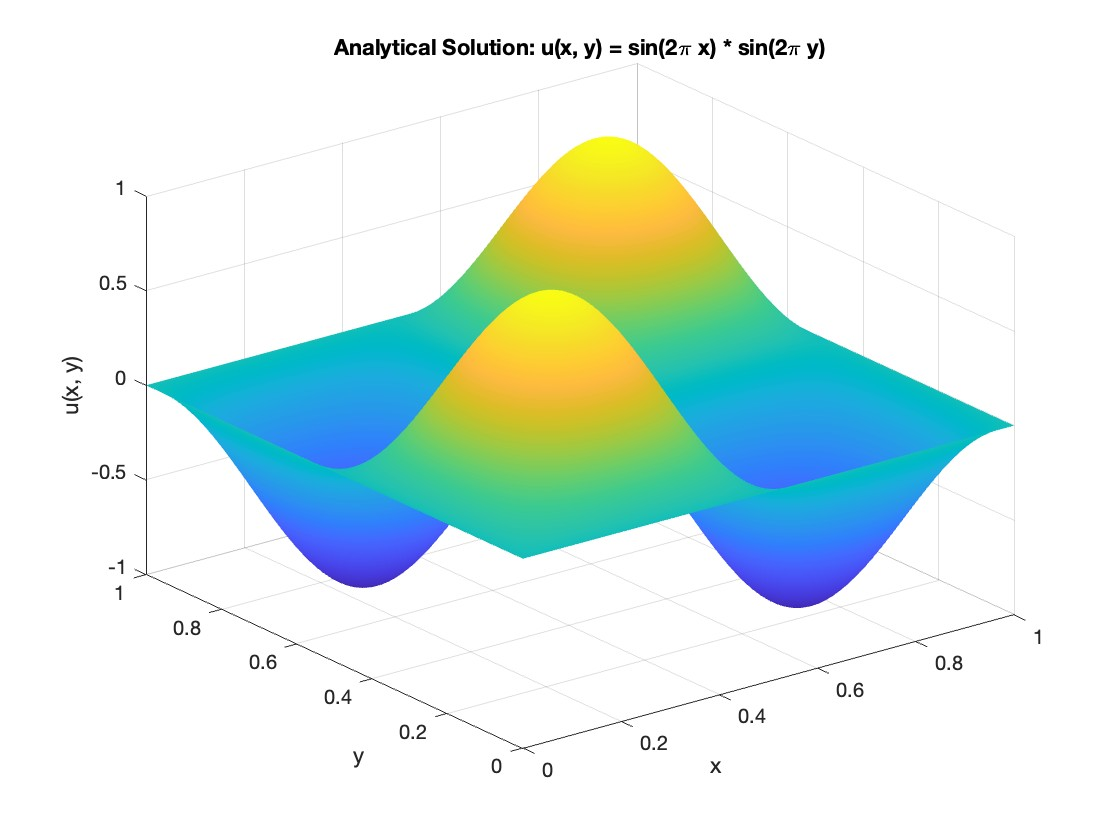}
        \caption{Analytic solution}
        \label{fig:cauchy}
    \end{minipage}
\end{figure}

\begin{figure}[h!]
    \centering
    \includegraphics[width=\linewidth,height=4cm,keepaspectratio]{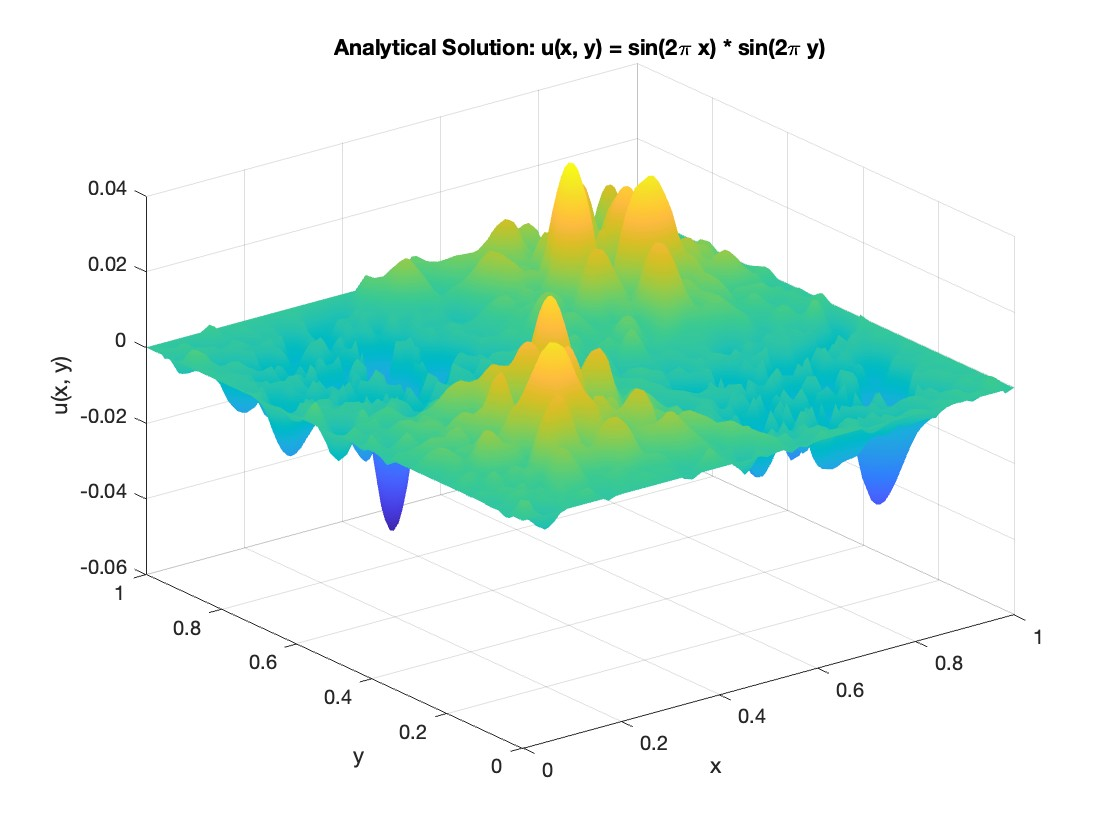}
    \caption{Pointwise Difference between two solutions}
\end{figure}

The $L^2$ error is computed as follows:
$\text{diff} = U_{\text{grid}} - F_{\text{grid}},$where $U_{\text{grid}}$ represents the predicted solution and $F_{\text{grid}}$ represents the analytical solution.

The $L^2$ error (mean squared error) is given by:
\begin{equation}
L^2_{\text{error}} = \sqrt{\frac{1}{n} \sum_{i=1}^{n} (\text{diff}_i)^2},
\end{equation}
where $n$ is the total number of grid points. The $L^2$ error is 0.0076886.

\FloatBarrier  

Next, we used a PINN model with a structure of size \([2, 200, 1]\), consisting of one hidden layer. The learning rate was set to 0.001 for the first 7000 epochs, and reduced to 0.0001 for the final 1000 epochs. The optimizer used was Adam. We compared the performance of two different activation functions: the tanh activation function and the Cauchy activation function over a total of 8000 training epochs. 

Our loss function consists of two parts: the first part represents the Mean Squared Error (MSE) of the residuals from the equation, while the second part accounts for the MSE of the boundary conditions:
\small
\begin{align*}
\text{Loss}&= \frac{1}{N_{\text{interior}}} \sum_{i=1}^{N_{\text{interior}}} \left( \frac{\partial^2 u(x_i)}{\partial x_1^2} + \frac{\partial^2 u(x_i)}{\partial x_2^2} - y_{\text{train}, i} \right)^2  +  \frac{1}{N_{\text{boundary}}} \sum_{j=1}^{N_{\text{boundary}}} \left( u(x_j) - u_{\text{boundary}, j} \right)^2. \\
\end{align*}

Since the loss values were initially large and decreased significantly over time, the overall loss curve did not clearly highlight the differences between the methods. To address this, we plotted the loss curve for the final 1000 epochs separately, making it evident that our method (Cauchy activation) holds a clear advantage over the original PINN.

\begin{figure}[h!]
    \centering
    \begin{minipage}[t]{0.48\textwidth}
        \centering
        \includegraphics[height=5cm, keepaspectratio]{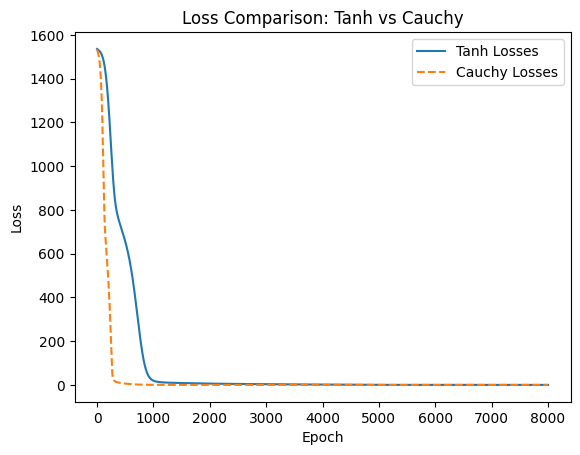}
        \caption{Overall loss curve for PINN using Tanh and Cauchy activations.}
        \label{fig:overall_loss_curve}
    \end{minipage}
    \hfill
    \begin{minipage}[t]{0.48\textwidth}
        \centering
        \includegraphics[height=5cm, keepaspectratio]{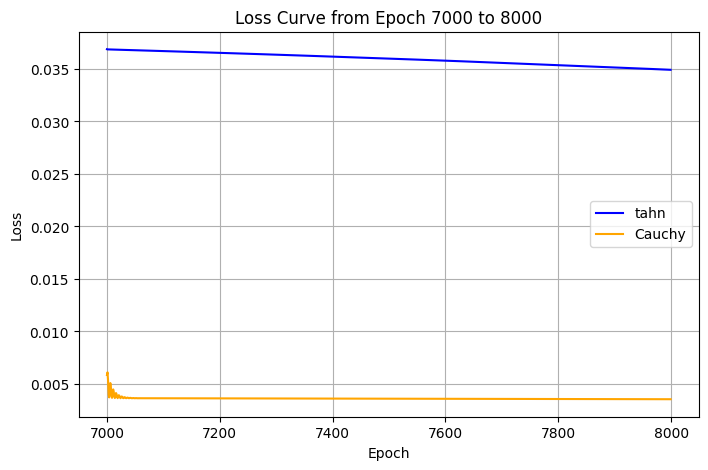}
        \caption{Zoomed-in loss curve for the final 1000 epochs.}
        \label{fig:zoomed_loss_curve}
    \end{minipage}
    \caption{Comparison of loss curves for PINN using Tanh and Cauchy activations. The Cauchy activation shows superior convergence speed and stability.}
    \label{fig:loss_curves}
\end{figure}
\FloatBarrier  

\begin{table}[h!]
    \centering
    \begin{tabular}{|c|c|}
        \hline
        \textbf{Activation Function} & \textbf{Training loss} \\
        \hline
        Tanh & 0.0349 \\
        \hline
        Cauchy & 0.00354 \\
        \hline
    \end{tabular}
    \caption{Comparison of activation functions and their respective training loss.}
\end{table}

\FloatBarrier  

The results clearly demonstrate that our XNet model significantly outperforms the standard activation functions, as seen in the graph where the Cauchy activation function consistently achieves a lower loss, showing superior effectiveness.   

 In addition to the one-hidden-layer model, we also tested a PINN model with two hidden layers, with a structure of size \([2, 20, 20, 1]\). The results for this two-hidden-layer model were similar to those of the one-hidden-layer model. Both models were trained using a learning rate of 0.001. After 8000 epochs, the tanh-based PINN achieved a training error of 0.119, while the XNet (Cauchy activation function) achieved a significantly lower training error of 0.0382.



\subsubsection{Burger's equation}
To validate the performance of our proposed Cauchy activation function, we included the Burger equation in our experiments. The Burger equation, commonly used in fluid dynamics, is defined as:

\[
\frac{\partial u}{\partial t} + u \frac{\partial u}{\partial x} = \nu \frac{\partial^2 u}{\partial x^2},
\]

where \(u(x, t)\) represents the velocity field, and \(\nu\) is the viscosity coefficient. 

In our implementation, we leveraged a Physics-Informed Neural Network (PINN) to solve the equation. The network approximates the solution \(u(x, t)\) and enforces the equation's physical constraints by minimizing the residual:

\[
f(x, t) = \frac{\partial u}{\partial t} + u \frac{\partial u}{\partial x} - \nu \frac{\partial^2 u}{\partial x^2}.
\]

\paragraph{Modifications for Experiments}

We simplified the network architecture by reducing its depth from 10 layers to 5 layers compared to the original design. This adjustment improved computational efficiency while maintaining the capacity to accurately model the solution.

In addition to this, we replaced traditional activation functions, such as \texttt{tanh}, with our custom Cauchy activation function in all layers. This modification was intended to leverage the rapid convergence and robust learning characteristics of the Cauchy activation function.

\paragraph{Results}

The results demonstrated a remarkable improvement in training efficiency:

\begin{itemize}
    \item With the \texttt{tanh} activation function, the network required \textbf{1000 epochs} to reduce the training loss to zero.
    \item With the Cauchy activation function, the training loss converged to zero within \textbf{20 epochs}, showcasing the superior convergence speed of our approach.
\end{itemize}

\begin{figure}[h!]
    \centering
    \includegraphics[width=0.75\linewidth]{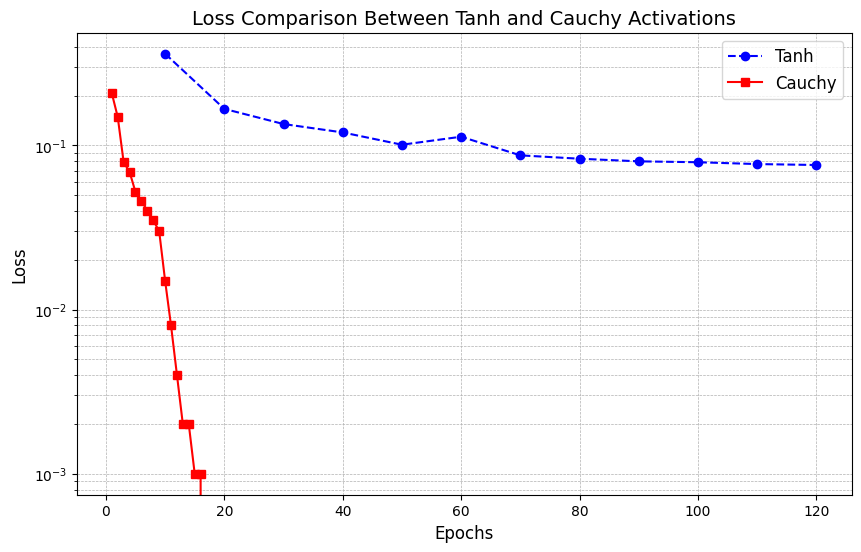}
    \caption{Training loss comparison for Burger's equation using \texttt{tanh} and Cauchy activation functions.}
    \label{fig:loss_burger}
\end{figure}

These findings validate that the Cauchy activation function is particularly well-suited for physics-informed neural networks, where rapid convergence and efficient training are critical.

\subsubsection{High Dimensional PDE }

In this section we test the XNet solver in the case of an 100-dimensional AllenCahn PDE with a cubic nonlinearity.

The Allen-Cahn equation is a reaction-diffusion equation that arises in physics, serving as a prototype for the modeling of phase separation and order-disorder transitions (see, e.g., \cite{Emmerich2003}). This equation is defined as:
\[
\frac{\partial u}{\partial t}(t, x) + u(t, x) - [u(t, x)]^3 + \left(\Delta_x u\right)(t, x) = 0,
\]
with the solution \( u \) satisfying for all \( t \in [0, T) \), \( x \in \mathbb{R}^d \):
\[
u(T, x) = g(x).
\]

Assume for all \( s, t \in [0, T] \), \( x, w \in \mathbb{R}^d \), \( y \in \mathbb{R} \), \( z \in \mathbb{R}^{1 \times d} \), \( m \in \mathbb{N} \) that \( d = 100 \), \( T = \frac{3}{10} \), \( \mu(t, x) = 0 \), \( \sigma(t, x)w = \sqrt{2}w \), and \( \Upsilon(s, t, x, w) = x + \sqrt{2}w \). The reaction term is defined as 
\[
f(t, x, y, z) = y - y^3,
\]
capturing the double-well potential of the Allen-Cahn equation, where the two minima at \( y = -1 \) and \( y = 1 \) represent stable equilibrium states. 

The terminal condition 
\[
g(x) = \left[2 + \frac{2}{5}\|x\|_{\mathbb{R}^d}^2\right]^{-1}
\]
ensures smooth decay for large \(\|x\|\), aligning with the expected physical behavior in bounded domains. These assumptions simplify the Allen-Cahn equation and provide a well-posed high-dimensional problem to evaluate the performance of the proposed method.

In our study, we employed a model identical to the one discussed in the paper \cite{E2017}. We simplified the original model by reducing the multilayer perceptron (MLP) component to a single layer, effectively halving the parameter count. Then, we substituted the activation function and evaluated performance differences. We set up an Adam optimizer with a learning rate of 0.005. The comparison model configuration remained the same as described in the original paper \cite{E2017} to ensure a fair evaluation.

\begin{table}[h]
\centering
\caption{Comparison of Training Loss between Cauchy and ReLU}
\label{tab:loss_comparison}
\begin{tabular}{|c|c|c|c|}
\hline
\textbf{Step} & \textbf{Cauchy Loss} & \textbf{ReLU Loss} & \textbf{Factor of Reduction} \\ \hline
0    & $1.5698 \times 10^{-1}$ & $1.5637 \times 10^{-1}$ & $\approx 1.00$ \\ \hline
100  & $2.9323 \times 10^{-3}$ & $9.8792 \times 10^{-2}$ & $\approx 33.70$ \\ \hline
200  & $3.3228 \times 10^{-3}$ & $7.8492 \times 10^{-2}$ & $\approx 23.63$ \\ \hline
500  & $2.9574 \times 10^{-3}$ & $3.7295 \times 10^{-2}$ & $\approx 12.61$ \\ \hline
1000 & $2.9667 \times 10^{-3}$ & $1.1308 \times 10^{-2}$ & $\approx 3.81$ \\ \hline
2000 & $2.3857 \times 10^{-3}$ & $4.9802 \times 10^{-3}$ & $\approx 2.09$ \\ \hline
\end{tabular}
\end{table}

\FloatBarrier


\begin{figure}[h!]
    \centering
        \includegraphics[width=\linewidth,height=4cm,keepaspectratio]{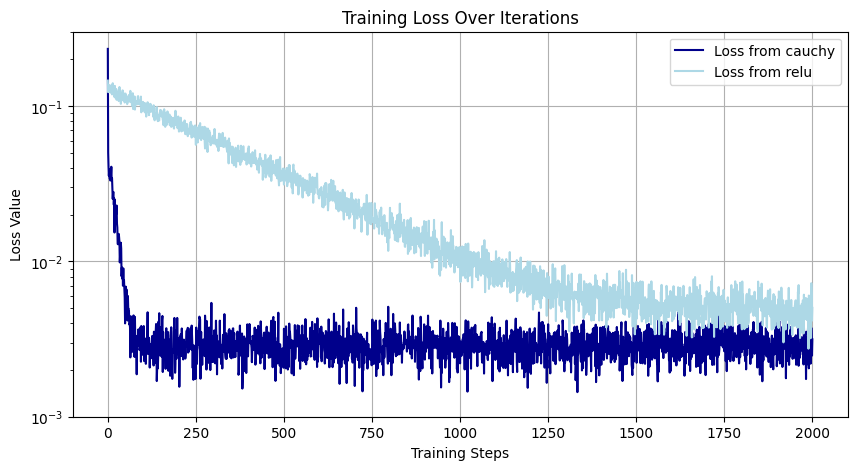} 
        \caption{loss curve of Allen Cahn }
        \label{fig:100}
      \end{figure}

\FloatBarrier

\appendix

\section{ Proofs for Theorem 2}


\noindent {\bf Proof:} The proof is based on Stone-Weierstrauss
Theorem: the set of all polynomials is dense in $C(B, \R)$ for any
compact subset $B \subset \R^N$. Since any polynomial is a linear
combination of monomials, we just need to show that linear
combinations of functions in $\Phi^N$ can approximate any monomial.

Since $\Phi$ has the approximation property, it therefore can approximate
$x^k$ for any integer $k$, this implies that $\Phi^N$ can approximate any
function of the form
$$(a_1x_1 + a_2x_2 + \ldots + a_Nx_N)^k, \; \mbox{ for } k\in \N,\;
(a_1, \ldots, a_N) \in \R^N.$$ For any fixed integer $k$, the
multinomial expansion of the above function is a linear combination of
$C(k+N-1, N-1)$ monomial terms such as
$$x_1^{k_1} x_2^{k_2} \cdots x_N^{k_N}, \; \; k_1 + k_2 + \ldots, k_N
= k.$$
Here $C(k+N-1, N-1)$ is combinatorial number $k+N-1$ choosing $N-1$. We claim that the converse is also true, i.e., every such monomial can be written as a linear
combination of functions of the form
$$(a_1x_1 + a_2x_2 + \ldots + a_Nx_N)^k.$$
Our theorem will directly follow from this claim.

We prove the claim by
induction on $N$.

\

For $N=1$, this is obviously true.

\

For $N=2$,
\be (x_1+ 0 x_2)^k &=& x_1^k \nn \\
(x_1 + 1x_2)^k &=& x_1^k + C(k, 1) x_1^{k-1}x_2 + \ldots + C(k, k-1)
x_1 x_2^{k-1} + x_2^k \nn \\
(x_1 + 2x_2)^k &=& x_1^k + C(k, 1) 2x_1^{k-1}x_2 + \ldots + C(k, k-1)
2^{k-1}x_1 x_2^{k-1} + 2^kx_2^k \nn \\
(x_1 + 3x_2)^k &=& x_1^k + C(k, 1) 3x_1^{k-1}x_2 + \ldots + C(k, k-1)
3^{k-1}x_1 x_2^{k-1} + 3^kx_2^k \nn \\
&...& \nn \\
(x_1 + kx_2)^k &=& x_1^k + C(k, 1) kx_1^{k-1}x_2 + \ldots + C(k, k-1)
k^{k-1}x_1 x_2^{k-1} + k^kx_2^k \nn \ee This is a system of equations for monomials of the form
$x_1^j x_2^{k-j}$.  The coefficient matrix is obviously
nonsingular,  therefore each $x_1^j x_2^{k-j}$ can be solved as a linear
combination of $(x_1 + jx_2)^k$, with $j=0, 1, \ldots, k$.

\

For $N=3$, fix any real number $l$, we can expand $((x_1+ lx_2) + j x_3)^k$ for
$j =0, 1, \ldots , k$ into the product of the form
$$(x_1 + l x_2)^{k-k_3} x_3^{k_3}, \; \mbox{ for } k_3=1,...,k-1.$$
Same as the
previous case with $N=2$, each of the above terms can be expressed as
a linear combination of

$$((x_1+ lx_2) + j x_3)^k, \; \mbox{ for
 }\; j =0, 1, \ldots, k.$$

Now for any fixed $k_3$, we choose $l = 0, 1, \ldots, k -k_3$ and expand
$$(x_1 + l x_2)^{k-k_3} x_3^{k_3}.$$
Similar to the case with $N=2$, we now obtain $k-k_3$ independent
equations. From these equations, we can solve each monomial 
$$x_1^{k_1}x_2^{k_2}x_3^{k_3}, \; \mbox{ for } k_1 + k_2 + k_3 =k$$
as a linear combination of function of the form $(x_1 + lx_2 +
jx_3)^k$, with $j=0, 1, \ldots, k$, $l = 0, 1, \ldots, k -k_3$. 

\

For any $N >3$, we continue this process. Any monomial of the form
$$x_1^{k_1} x_2^{k_2} \cdots x_N^{k_N}, \; \; k_1 + k_2 + \ldots, k_N
= k$$
can be written as a linear
combination of functions of the form
$$(a_1x_1 + a_2x_2 + \ldots + a_Nx_N)^k$$
with $a_1=1$; $a_2 = 0, \ldots, k_1$; $a_3 =0, \ldots, k_1 + k_2$;
$\ldots$; and $a_n = 0, 1, \ldots, k$.

This proves the theorem. \qed


\begin{thebibliography}{99}


\bibitem{ZLX24}
Hongkun Zhang, Xin Li, and Zhihong Xia.
\textit{CauchyNet: Utilizing Complex Activation Functions for Enhanced Time-Series Forecasting and Data Imputation}.
Submitted for publication, 2024.




\bibitem{Hirose2012}
Akira Hirose.
\textit{Complex-Valued Neural Networks}.
Springer Science \& Business Media, volume 400, 2012.

\bibitem{Goodfellow2016}
Ian Goodfellow, Yoshua Bengio, and Aaron Courville.
\textit{Deep Learning}.
MIT Press, 2016.



\bibitem{Rumelhart1986}
David E. Rumelhart, Geoffrey E. Hinton, and Ronald J. Williams.
\textit{Learning representations by back-propagating errors}.
Nature, volume 323, number 6088, pages 533--536, 1986. DOI: 10.1038/323533a0.

\bibitem{DeVore1989}
Ronald A. DeVore and Ralph Howard and Charles Micchelli.
\textit{Optimal nonlinear approximation}.
Manuscripta Mathematica, volume 63, number 4, pages 469--478, 1989.

\bibitem{Mhaskar1996}
Hrushikesh N. Mhaskar.
\textit{Neural Networks for Optimal Approximation of Smooth and Analytic Functions}.
Neural Computation, volume 8, number 1, pages 164--177, 1996.

\bibitem{LeCun2015}
Yann LeCun, Yoshua Bengio, and Geoffrey Hinton.
\textit{Deep learning}.
Nature, volume 521, number 7553, pages 436--444, 2015, DOI: 10.1038/nature14539.


\bibitem{Li2024}
Xin Li, Zhihong Jeff Xia, and Xiaotao Zheng.
\textit{Model Comparisons: XNet Outperforms KAN}.
arXiv preprint arXiv:2410.02033, 2024.


\bibitem{E2017}
Jiequn Han, Arnulf Jentzen, and Weinan E.
\textit{Solving high-dimensional partial differential equations using deep learning}.
Proceedings of the National Academy of Sciences, volume 115, number 34, pages 8505--8510, 2018.

\bibitem{Emmerich2003}
H. Emmerich.
\textit{The Diffuse Interface Approach in Materials Science: Thermodynamic Concepts and Applications of Phase-Field Models}.
Springer Science \& Business Media, volume 73, 2003.



\bibitem{Senior2020}
Andrew W. Senior, Richard Evans, John Jumper, James Kirkpatrick, Laurent Sifre, Tim Green, and Demis Hassabis.
\textit{Improved protein structure prediction using potentials from deep learning}.
Nature, volume 577, number 7792, pages 706--710, 2020, DOI: 10.1038/s41586-019-1923-7.

\bibitem{Jumper2021}
John Jumper, Richard Evans, Alexander Pritzel, Tim Green, Michael Figurnov, Kathryn Tunyasuvunakool, and Demis Hassabis.
\textit{Highly accurate protein structure prediction with AlphaFold}.
Nature, volume 596, number 7873, pages 583--589, 2021, DOI: 10.1038/s41586-021-03819-2.

\bibitem{Lee2022}
Chi Yan Lee, Hideyuki Hasegawa, and Shangce Gao.
\textit{Complex-Valued Neural Networks: A Comprehensive Survey}.
IEEE/CAA Journal of Automatica Sinica, volume 9, number 8, pages 1406--1426, 2022.

\bibitem{Barrachina2023}
Jose Agustin Barrachina, Chengfang Ren, Gilles Vieillard, Christele Morisseau, and Jean-Philippe Ovarlez.
\textit{Theory and Implementation of Complex-Valued Neural Networks}.
arXiv preprint arXiv:2302.08286, 2023.

\bibitem{Reichstein2019}
Markus Reichstein, Gustau Camps-Valls, Bjorn Stevens, Martin Jung, Joachim Denzler, Nuno Carvalhais, and Prabhat.
\textit{Deep learning and process understanding for data-driven Earth system science}.
Nature, volume 566, number 7743, pages 195--204, 2019, DOI: 10.1038/s41586-019-0912-1.

\bibitem{Kunc2024}
Vladim{\'\i}r Kunc and Ji{\v{r}}{\'\i} Kl{\'e}ma.
\textit{Three Decades of Activations: A Comprehensive Survey of 400 Activation Functions for Neural Networks}.
arXiv, 2024, URL: https://ar5iv.org/abs/2402.09092.

\bibitem{ModSwish2024}
Heena Kalim, Anuradha Chug, and Amit Singh.
\textit{modSwish: a new activation function for neural network}.
Evolutionary Intelligence, volume 17, pages 1-11, 2024, DOI: 10.1007/s12065-024-00908-9.

\bibitem{Dubey2022}
S. R. Dubey, S. K. Singh, and B. B. Chaudhuri.
\textit{Activation functions in deep learning: A comprehensive survey and benchmark}.
Neurocomputing, volume 503, pages 92--108, 2022, DOI: 10.1016/j.neucom.2022.06.111.

\bibitem{Bingham2022}
G. Bingham and R. Miikkulainen.
\textit{Discovering Parametric Activation Functions}.
Neural Networks, volume 148, pages 48--65, 2022, DOI: 10.1016/j.neunet.2022.01.001.

\bibitem{Ramachandran2017}
P. Ramachandran, B. Zoph, and Q. V. Le.
\textit{Searching for Activation Functions}.
2017, arXiv: 1710.05941, DOI: 10.48550/ARXIV.1710.05941, URL: https://arxiv.org/abs/1710.05941.



\bibitem{Knezevic2023}
K. Knezevic, J. Fulir, D. Jakobovic, S. Picek, and M. Durasevic.
\textit{NeuroSCA: Evolving Activation Functions for Side-Channel Analysis}.
IEEE Access, volume 11, pages 284--299, 2023, DOI: 10.1109/access.2022.3232064.

\bibitem{Lu2021}
L. Lu, X. Meng, Z. Mao, and G. E. Karniadakis.
\textit{DeepXDE: A Deep Learning Library for Solving Differential Equations}.
SIAM Review, volume 63, number 1, pages 208--228, 2021, DOI: 10.1137/19M1274067.

\bibitem{Sirignano2018}
J. Sirignano and K. Spiliopoulos.
\textit{DGM: A Deep Learning Algorithm for Solving Partial Differential Equations}.
Journal of Computational Physics, volume 375, pages 1339--1364, 2018, DOI: 10.1016/j.jcp.2018.08.029.

\bibitem{Raissi2019}
M. Raissi, P. Perdikaris, and G. E. Karniadakis.
\textit{Physics-Informed Neural Networks: A Deep Learning Framework for Solving Forward and Inverse Problems Involving Nonlinear Partial Differential Equations}.
Journal of Computational Physics, volume 378, pages 686--707, 2019, DOI: 10.1016/j.jcp.2018.10.045.

\bibitem{Jin2021}
P. Jin, L. Lu, G. Pang, Z. Zhang, and G. E. Karniadakis.
\textit{Learning Nonlinear Operators via DeepONet Based on the Universal Approximation Theorem of Operators}.
Nature Machine Intelligence, volume 3, number 3, pages 218--229, 2021, DOI: 10.1038/s42256-021-00302-5.

\bibitem{Transolver2024}
H. Wu, H. Luo, H. Wang, J. Wang, and M. Long.
\textit{Transolver: A Fast Transformer Solver for PDEs on General Geometries}.
arXiv preprint arXiv:2402.02366, 2024.


\bibitem{yarotsky2017error}
D. Yarotsky.
\textit{Error bounds for approximations with deep ReLU networks}.
Neural Networks, 94:103--114, 2017.

\bibitem{e2018exponential}
W. E, Q. Wang.
\textit{A priori estimates and {P}och{H}ammer-{C}hree expansions for deep neural networks}.
Communications in Mathematical Sciences, 16(8):2349--2383, 2018.

\bibitem{PINNsFormer2023}
Z. Zhao, X. Ding, and B. Aditya Prakash.
\textit{PINNsFormer: A Transformer-Based Framework For Physics-Informed Neural Networks}.
arXiv preprint arXiv:2307.11833, 2023.

\bibitem{le2022deep}
N. Le, V. Rathour, K. Yamazaki, K. Luu, and M. Savvides.
\textit{Deep reinforcement learning in computer vision: a comprehensive survey}.
Artificial Intelligence Review, pages 1--87, 2022, Springer.

\bibitem{kaur2023comprehensive}
R. Kaur and S. Singh.
\textit{A comprehensive review of object detection with deep learning}.
Digital Signal Processing, volume 132, pages 103812, 2023, Elsevier.






\bibitem{wu2020visual}
B. Wu, C. Xu, X. Dai, A. Wan, P. Zhang, Z. Yan, M. Tomizuka, J. Gonzalez, K. Keutzer, and P. Vajda.
\textit{Visual transformers: Token-based image representation and processing for computer vision}.
arXiv preprint arXiv:2006.03677, 2020.

\bibitem{fu2024featup}
S. Fu, M. Hamilton, L. Brandt, A. Feldman, Z. Zhang, and W. T. Freeman.
\textit{Featup: A model-agnostic framework for features at any resolution}.
arXiv preprint arXiv:2403.10516, 2024.

\bibitem{Tripathi2021}
R. P. Tripathi, M. Tiwari, A. Dhawan, A. Sharma, and S. K. Jha.
\textit{A Survey on Efficient Realization of Activation Functions of Artificial Neural Network}.
2021 Asian Conference on Innovation in Technology (ASIANCON), IEEE, 2021. DOI: 10.1109/asiancon51346.2021.9544754.


\bibitem{Jagtap2020a}
A. D. Jagtap, K. Kawaguchi, and G. E. Karniadakis.
\textit{Adaptive Activation Functions Accelerate Convergence in Deep and Physics-Informed Neural Networks}.
Journal of Computational Physics, vol. 404, 2020, pp. 109136. DOI: 10.1016/j.jcp.2019.109136.

\bibitem{Jagtap2020b}
A. D. Jagtap and G. E. Karniadakis.
\textit{Extended Physics-Informed Neural Networks (XPINNs): A Generalized Space-Time Domain Decomposition Based Deep Learning Framework for Nonlinear Partial Differential Equations}.
Communications in Computational Physics, vol. 28, no. 5, 2020, pp. 2002-2041. DOI: 10.4208/cicp.OA-2020-0160.

\bibitem{Jagtap2022}
A. D. Jagtap, E. Kharazmi, and G. E. Karniadakis.
\textit{Locally Adaptive Activation Functions with Applications to Deep and Physics-Informed Neural Networks}.
Journal of Scientific Computing, vol. 92, 2022, pp. 80. DOI: 10.1007/s10915-022-01759-8.

\bibitem{Jagtap2022Survey}
A. D. Jagtap and G. E. Karniadakis,
\textit{How Important Are Activation Functions in Regression and Classification? A Survey, Performance Comparison, and Future Directions},
arXiv:2209.02681, 2022.



\bibitem{Begion2016}
M. Arjovsky, A. Shah, and Y. Bengio.
\textit{Unitary Evolution Recurrent Neural Networks}.
Proceedings of the 33rd International Conference on Machine Learning - Volume 48, JMLR.org, 2016, pages 1120--1128.

\bibitem{Jarrett2009}
K. Jarrett, K. Kavukcuoglu, M. Ranzato, and Y. LeCun.
\textit{What is the best multi-stage architecture for object recognition?}
Proceedings of the IEEE International Conference on Computer Vision (ICCV), IEEE, 2009, Kyoto, Japan, pages 2146--2153. DOI: 10.1109/ICCV.2009.5459469.

\bibitem{Hinton2010}
V. Nair and G. E. Hinton.
\textit{Rectified linear units improve restricted boltzmann machines}.
Proceedings of the 27th international conference on machine learning (ICML-10), 2010, pages 807--814.

\bibitem{Maas2013}
A. L. Maas, A. Y. Hannun, and A. Y. Ng.
\textit{Rectifier Nonlinearities Improve Neural Network Acoustic Models}.
Proceedings of the 30th International Conference on Machine Learning (ICML), volume 30, 2013, page 3.

\bibitem{He2015}
K. He, X. Zhang, S. Ren, and J. Sun.
\textit{Delving Deep into Rectifiers: Surpassing Human-Level Performance on ImageNet Classification}.
Proceedings of the IEEE International Conference on Computer Vision (ICCV), 2015, pages 1026--1034.

\bibitem{Klambauer2017}
G. Klambauer, T. Unterthiner, A. Mayr, and S. Hochreiter.
\textit{Self-Normalizing Neural Networks}.
Advances in Neural Information Processing Systems (NeurIPS), 2017, pages 971--980.

\bibitem{Telgarsky2017}
M. Telgarsky.
\textit{Neural networks and rational functions}.
Proceedings of the 34th International Conference on Machine Learning (ICML), volume 70, 2017, pages 3387--3393.

\bibitem{Begion2018}
Y. Begion and others.
\textit{Deep Complex Networks}.
Proceedings of the 6th International Conference on Learning Representations, 2018, Poster session.

\bibitem{Yeats2021}
E. C. Yeats, Y. Chen, and H. Li.
\textit{Improving Gradient Regularization using Complex-Valued Neural Networks}.
Proceedings of the 38th International Conference on Machine Learning, volume 139, PMLR, 2021, pages 11953--11963.

\bibitem{Boulle2020}
N. Boull{\'{e}}, Y. Nakatsukasa, and A. Townsend.
\textit{Rational Neural Networks}.
Advances in Neural Information Processing Systems, volume 33, 2020, pages 14243--14253.


\end{thebibliography}
\end{document}